\pdfoutput=1

\documentclass[11pt]{article}

\usepackage[]{ACL2023}

\usepackage{times}
\usepackage{latexsym}
\usepackage{placeins}
\usepackage{afterpage}
\usepackage{graphicx}
\usepackage{subcaption}
\usepackage{makecell}
\usepackage{float}
\usepackage{tabularx}
\usepackage{booktabs}
\usepackage{ragged2e}
\usepackage{stfloats}
\usepackage[most]{tcolorbox}
\usepackage[table]{xcolor}

\usepackage{colortbl}
\definecolor{quiettgrey}{RGB}{240, 240, 240}
\definecolor{quiettblue}{RGB}{219, 234, 254}
\definecolor{quiettgreen}{RGB}{220, 245, 220}
\definecolor{quiettyellow}{RGB}{255, 251, 219}
\tcbuselibrary{breakable,skins}

\usepackage{titlesec}

\titlespacing*{\section}
  {0pt}        
  {8pt}        
  {4pt}        

\titlespacing*{\subsection}
  {0pt}
  {6pt}
  {3pt}

\titlespacing*{\subsubsection}
  {0pt}
  {5pt}
  {2pt}

\titlespacing*{\paragraph}
  {0pt}
  {4pt}
  {0.5em}      

\usepackage{listings}
\usepackage[T1]{fontenc}

\usepackage[utf8]{inputenc}
\usepackage{amsmath}
\usepackage{algorithm}       
\usepackage{algpseudocode}   
\usepackage{float}
\floatstyle{ruled}
\restylefloat{algorithm}

\usepackage{booktabs}
\usepackage{multirow}
\usepackage{array}

\usepackage{microtype}

\usepackage{inconsolata}

\title{\textsc{QuIeTT}: Query-Independent Table Transformation for Robust Reasoning}

\author{
\textbf{Gaurav Najpande} \qquad
\textbf{Tampu Ravi Kumar}\textsuperscript{*} \qquad
\textbf{Manan Roy Choudhury}\textsuperscript{*} \\
\textbf{Neha Valeti}\textsuperscript{*} \qquad
\textbf{Yanjie Fu} \qquad
\textbf{Vivek Gupta} \\
Arizona State University \\
\texttt{\{gnajpand,traviku2,mroycho1,nvaleti1,yanjie.fu,vgupt140\}@asu.edu} \\
}

\begin{document}
\maketitle
\begin{abstract}
Real-world tables often contain schema inconsistencies, 
heterogeneous value formats, and implicit relational 
structures that degrade table reasoning and question 
answering. Existing approaches address these issues at 
query time, repeating normalization and restructuring for 
every new question and producing representations that do 
not generalize to unseen queries. We introduce 
\textbf{\textsc{QuIeTT}}, a \textit{transform-first, 
query-later} framework that converts each raw table into 
a single SQL-ready canonical representation before any 
evaluation query is observed, under three constraints: 
no evaluation query access, information preservation, and 
single-table reuse. \textsc{QuIeTT} operates through a 
three-stage pipeline: issue probing, where synthetic 
queries surface structural deficiencies such as ambiguous 
schemas and heterogeneous formats; plan generation; and 
structured execution. Experiments on four benchmarks 
across five model families show consistent improvements 
over strong baselines in both direct prompting and 
agentic reasoning paradigms, with particularly strong 
gains on a manually annotated challenge set of 
structurally diverse unseen questions.

\let\thefootnote\relax\footnotetext{Code and data available at \url{https://anonymous.4open.science/r/QuIeTT-2B73/}}
\end{abstract}

\section{Introduction}

Table based reasoning is a central component of many natural language understanding tasks, including question answering, information retrieval and decision support~\cite{zhu2021retrievingreadingcomprehensivesurvey, herzig-etal-2020-tapas}, yet real-world tables rarely arrive in a form suitable for natural-language or symbolic
querying. For example, result columns embed outcome and score in a single string (\texttt{"W 24-17"}), columns mix scales and currencies (\texttt{"\$1.2M"}, \texttt{"EUR 850K"},
\texttt{"1,500,000"}), symbols stand in for missing values
(\texttt{"†"}, \texttt{"-"}), and hierarchical headers collapse
ambiguously when flattened~\cite{cheng-etal-2022-hitab}.

\begin{figure}[t]
\centering
\includegraphics[width=0.95\columnwidth]{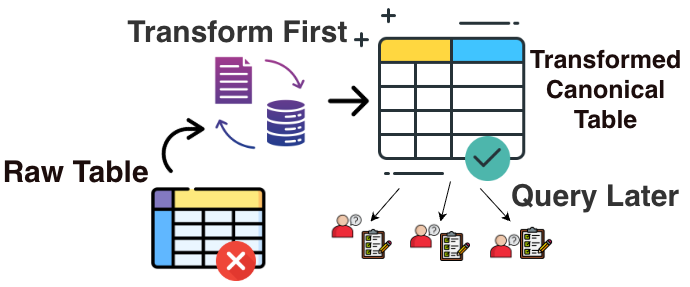}
\vspace{-0.75em}
\caption{\small \textsc{QuIeTT} transforms a raw table into a single SQL-ready canonical representation before any evaluation query is observed. The transformed table is reused across all queries without modification.}
\label{fig:teaser}
\vspace{-1.0em}
\end{figure}


Normalization methods such as NormTab~\cite{nahid2025normtab} and 
E5~\cite{zhang2024e5} reduce surface-form variability by 
standardizing values and applying lightweight structural fixes, but 
leave schema-level issues including heterogeneous layouts, implicit 
relations, and undecomposed attributes to downstream reasoning. 
Because these methods assume mostly fixed schemas, they cannot 
resolve the structural deficiencies that degrade reasoning on 
real-world tables.

Reasoning-coupled methods integrate transformation directly into 
the query pipeline. Chain-of-Table~\cite{wang2024chainoftable} 
and TableReasoner~\cite{xiong2025tablereasoneradvancingtablereasoning} 
perform query-driven schema selection and iterative restructuring; 
Binder~\cite{Binder} and Dater~\cite{10.1145/3539618.3591708} condition 
execution and decomposition on the observed question. These methods 
improve accuracy on seen queries but couple table representation to 
the query at hand: the same table yields different transformed views 
for different questions, transformation work repeats for every new 
query, and representation quality cannot be evaluated independently 
of reasoning behavior. In every case, reuse across questions, 
models, and tasks is structurally precluded.

Standard data engineering offers a useful analogy. In the widely adopted medallion architecture, data engineers prepare analytics-ready representations before any downstream query is posed~\cite{10.1145/2882903.2912574}, and that single representation serves diverse analytical workloads without query-specific adaptation. Table QA lacks an equivalent. \textsc{QuIeTT} draws on this intuition, motivated by the medallion architecture's principle of preparing analytics-ready representations before queries arrive. However, applying this principle to open-domain table QA introduces challenges that ETL does not face. ETL assumes a known target schema and a fixed query workload. In open-domain table QA, tables arrive with unknown structure and no future queries are observed at transformation time. \textsc{QuIeTT} addresses this by treating schema discovery as intrinsic to transformation: it derives the canonical representation from the raw table's own structural properties through issue probing. The result is a framework evaluated by a criterion that ETL systems never impose: a single canonical representation, constructed without query access and without a predefined schema, must improve downstream QA across diverse datasets, model families, and reasoning strategies.

We introduce \textsc{QuIeTT}, a \emph{transform-first,
query-later} framework that converts each raw table into a single SQL-ready canonical representation before any evaluation query is observed. \textsc{QuIeTT} operates under three explicit constraints: it accesses no downstream evaluation queries during transformation, it preserves all original table content so that nothing is discarded, and it produces exactly one canonical table reused across all queries. These constraints isolate the effect of table representation from model and prompt variation. When the same reasoning model and prompting strategy are held fixed, performance differences between raw and canonical tables reflect representation quality alone. \textsc{QuIeTT} targets the regime where tables are 
preprocessed once and reused across arbitrary future 
queries, complementing rather than replacing 
query-adaptive methods that condition transformation 
on observed queries.

\noindent\textbf{Our main contributions are:} \textbf{(1)} We introduce \textsc{QuIeTT}, a three-stage transform-first framework that converts raw tables into a single SQL-ready canonical representation before any evaluation query is observed, decoupling table preprocessing from query-time reasoning. \textbf{(2)} Across four benchmarks and five model families, \textsc{QuIeTT} consistently outperforms strong baselines by 4-8 F1 and 3-6 accuracy points across both direct prompting and agentic reasoning paradigms. \textbf{(3)} We construct a manually annotated challenge set of 2,500 questions over 529 tables (Fleiss'~$\kappa = 0.93$) targeting structural diversity absent from standard benchmarks. \textbf{(4)} Ablation studies show that issue probing is the primary driver of gains, with transformation quality mattering more than QA model capacity.

\begin{figure*}[t]
  \centering
  \includegraphics[width=1.0\linewidth]{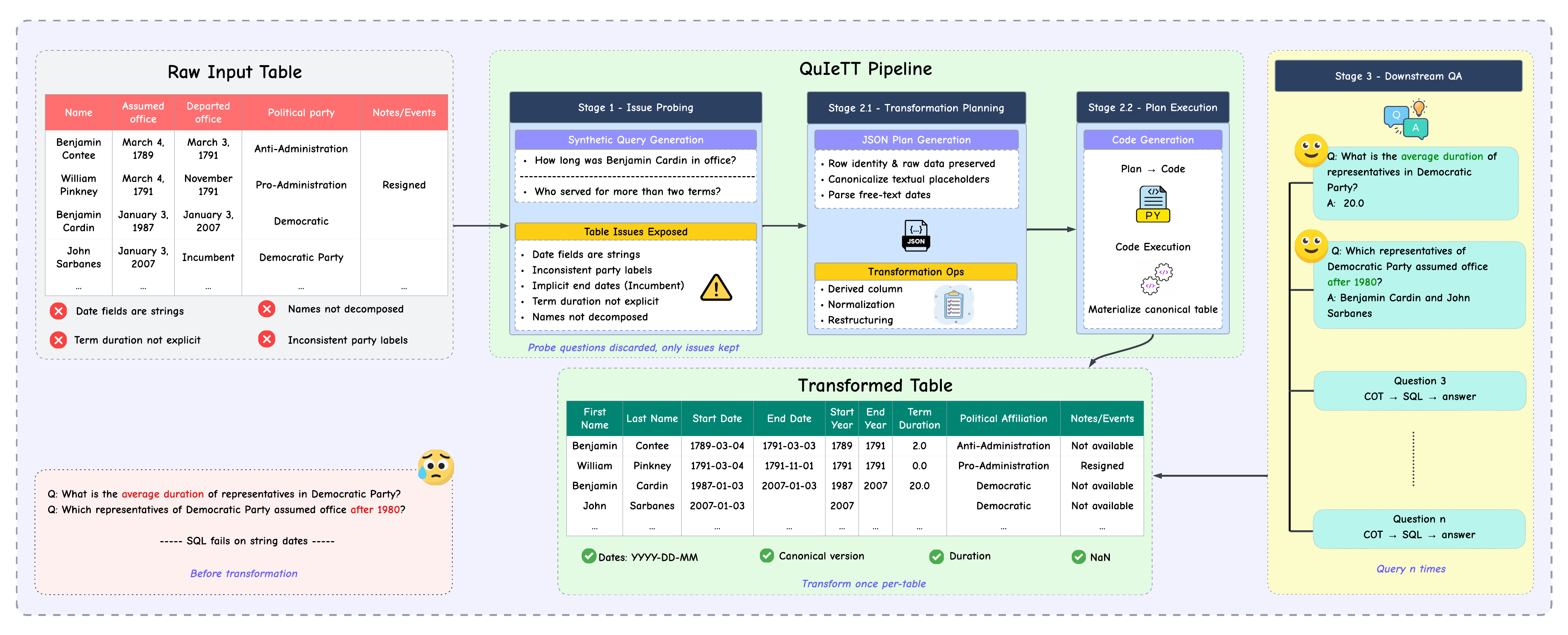}
  \vspace{-1.0em}
  \caption{\small \textsc{QuIeTT} framework for robust table reasoning.
  \textbf{Stage 1: Issue Probing.} Synthetic queries are generated
  from the raw table to expose structural issues such as inconsistent
  formats, implicit relations, missing derived fields, and
  undecomposed attributes.
  \textbf{Stage 2: Transformation.} Detected issues guide a
  transformation plan that applies normalization, parsing,
  restructuring, and derived column creation, executed via code to
  materialize a canonical, SQL-ready table.
  \textbf{Stage 3: QA.} The transformed table is reused across all
  downstream queries, enabling accurate and consistent reasoning
  without modifying the underlying model or prompting strategy.}
  \label{fig:pipeline_diagram}
  \vspace{-1.5em}
\end{figure*}

\section{\textsc{QuIeTT} Framework}
\label{sec:framework}

\subsection{Problem Setup}
\label{sec:problem_setup}
Let $T_{\text{raw}} \in \mathcal{T}$ denote a raw, semi-structured 
table and $\mathcal{Q}$ the space of possible downstream queries. 
We construct a transformation function $f : \mathcal{T} \rightarrow 
\mathcal{T}$ that maps $T_{\text{raw}}$ to a canonical, SQL-ready 
table $T_{\text{C}} = f(T_{\text{raw}})$, reused across all 
subsequent queries without modification, under three constraints:

\noindent\textbf{(a) No query access.} $f$ depends only on 
$T_{\text{raw}}$, not on any $q \in \mathcal{Q}$.

\noindent\textbf{(b) Information preservation.} All original content is grounded in $T_{\text{C}}$; $f$ 
reorganizes and normalizes rather than discards, preserving 
source-table provenance without dropping information.

\noindent\textbf{(c) Single-table reuse.} $T_{\text{C}}$ is produced 
exactly once and reused unchanged, so performance differences between 
$T_{\text{raw}}$ and $T_{\text{C}}$ reflect representation quality 
alone.

\noindent Given a fixed reasoning model $M$ and prompting strategy 
$S$, a query $q$ is answered as $\hat{a} = M(S(q, T_{\text{C}}))$, 
isolating the effect of table representation from model or prompt 
variation.

\subsection{Pipeline Overview}

\textsc{QuIeTT}
instantiates this setting as a three-stage pipeline (Algorithm~\ref{alg:quiett}):
(1)~probe the raw table to identify structural issues,
(2)~generate a transformation plan that resolves them, and
(3)~execute the plan to produce the canonical table $T_{\text{C}}$.

{
\vspace{-0.75em}
\begin{algorithm}[t]
\small
\ttfamily
\caption{\small \textsc{QuIeTT} Framework}
\label{alg:quiett}
\begin{algorithmic}
\State \textbf{Input:} raw table $T_{\mathrm{raw}}$
\State \textbf{Output:} canonical table $T_{\mathrm{C}}$,
    answers $\{\hat{a}\}$
\State \textbf{\textit{Stage 1}: Issue Probing}
\State $\mathcal{Q}_{\mathrm{probe}} \leftarrow$
    \textbf{GenerateSyntheticQueries}$(T_{\mathrm{raw}})$
\State $\mathcal{I} \leftarrow$
    \textbf{ExtractIssues}$(\mathcal{Q}_{\mathrm{probe}})$
    \Comment{probes discarded}
\State $\pi \leftarrow$ \textbf{GeneratePlan}$(T_{\mathrm{raw}},
    \mathcal{I})$
\State \textbf{Validate}$(\pi)$
    \Comment{format, completeness}
\State \textbf{\textit{Stage 2}: Transformation}
\State $T \leftarrow T_{\mathrm{raw}}$
\For{operation $o \in \pi$}
    \If{$o$ is deterministic}
        \State $T \leftarrow$ \textbf{ApplyOp}$(o, T)$
    \Else
        \State $T \leftarrow$
            \textbf{GenerateAndExecuteCode}$(o, T)$
    \EndIf
\EndFor
\State $T_{\mathrm{C}} \leftarrow T$
    \Comment{information preserved}
\State \textbf{\textit{Stage 3}: Downstream QA}
\For{query $q$}
    \State $r \leftarrow$ \textbf{Reason}$(q, T_{\mathrm{C}})$
        \Comment{CoT $\rightarrow$ SQL}
    \State $\hat{a} \leftarrow$ \textbf{Execute}$(r, T_{\mathrm{C}})$
\EndFor
\State \textbf{Return} $T_{\mathrm{C}}, \{\hat{a}\}$
\end{algorithmic}
\end{algorithm}
}
\vspace{-1.00em}

\vspace{1.50em}
\subsection{Issue Probing}

Issue probing generates a set of synthetic queries
$\mathcal{Q}_{\mathrm{probe}}(T_{\mathrm{raw}}) =
\{q^{(1)}_{\mathrm{syn}},\dots,q^{(M)}_{\mathrm{syn}}\}$
directly from $T_{\mathrm{raw}}$, where each query targets a
structural pattern that would hinder SQL or general-purpose querying,
such as ambiguous column schemas, heterogeneous value formats, type
inconsistencies, packed multi-value cells, or implicit relational
structure (see Appendix~\ref{app:issue_gen} for prompt design and
issue aggregation). Each probe query carries a \texttt{failure\_reason}
field that records why the raw table cannot answer it reliably;
grouping queries by identical failure reasons yields the issue set
$\mathcal{I}(T_{\mathrm{raw}}) = \{i_1, i_2, \dots, i_N\}$ that
drives transformation planning. The probe queries themselves are
discarded after issue extraction and later stages access only
$\mathcal{I}(T_{\mathrm{raw}})$, preserving constraint~(a). We
generate 12-20 probe queries per table; Appendix~\ref{sec:probe_analysis}
shows this budget achieves 98.5\% structural coverage.

\noindent\textbf{Handling hierarchical tables.}
Hierarchical tables encode information across implicit parent-child
levels rather than explicit columns~\cite{zhao-etal-2023-large},
which can cause issue probing to miss cross-level dependencies.
Before probing, we apply a deterministic flattening step that
collapses multi-level headers into composite column names by
concatenating parent and child labels with a separator
(e.g., \texttt{eastern ontario / french-language workers}),
as illustrated in Appendix~\ref{app:examples}. While this produces
a uniform columnar representation that issue probing can operate
over, the underlying semantics remain implicitly encoded in column
names rather than explicit fields. Issue probing then surfaces these
residual structural deficiencies, and the transformation plan
resolves them by decomposing composite headers into atomic attributes
such as region and worker type.

\subsection{Transformation Planning}

A transformation plan is an ordered sequence
$\pi=\langle o_1,\dots,o_K\rangle$, where the plan length $K$ and the
specific operations are determined dynamically from the issue set
$\mathcal{I}(T_{\mathrm{raw}}) = \{i_1, i_2, \dots, i_N\}$ detected
in Stage~1. Each operation $o_i$ declares its input columns, output
columns, and dependencies, yielding an auditable transformation graph.
Collectively, $\pi$ induces an explicit target schema
$\mathcal{S}_{\pi}$ that fixes the column set, names, types, and
derived fields of the transformed table before any execution begins.
The plan is produced by a function
$\mathcal{P}:\mathcal{T}\times\mathcal{I}\rightarrow\Pi$, giving
$\pi=\mathcal{P}(T_{\mathrm{raw}},\mathcal{I}(T_{\mathrm{raw}}))$.
Constraining operation semantics while allowing variable plan length
and composition lets \textsc{QuIeTT} produce reproducible plans while
limiting hallucinated or unsupported transformations.

\subsection{Plan Execution}
The plan $\pi$ is executed by a structured engine that separates
symbolic planning from code-level implementation, following the design
principle of~\citet{khoja-etal-2025-weaver}. Given $\pi$ and
$T_{\mathrm{raw}}$, the executor processes each operation
$o_i \in \pi$ according to its declared inputs, outputs, and
dependencies, producing the canonical table
$T_{\mathrm{C}} = E(\pi, T_{\mathrm{raw}})$.

Execution follows two paths depending on the operation type.
Operations with registered implementations (see 
Appendix~\ref{app:operators}) are applied directly through a fixed operator
library. Operations requiring table-specific logic are handled by
LLM-generated Python code executed against the current table state.
In both paths, each step is restricted to its declared inputs and
outputs in $\pi$. Steps that fail validation trigger a retry: the validator appends 
the error message to the prompt and re-queries the model up to two 
additional times. If all attempts fail, the original column values 
are preserved as a fallback and execution continues with the 
next plan step, bounding the effect of any variability in 
LLM-generated steps to the plan structure.

The executor enforces schema conformance throughout, so the final
table conforms to $\mathcal{S}_{\pi}$ regardless of which path
individual steps take. Whenever a transformation changes the original
surface form of a value, such as parsing a date or splitting a packed
cell into separate columns, the original information is reorganized
into the new representation rather than discarded. Columns that
require no transformation are carried over to $T_{\mathrm{C}}$
unchanged (see Appendix~\ref{app:grounding_analysis}). Because
$T_{\mathrm{C}}$ is produced once and cached, the pre-processing cost
is amortized across all downstream queries
(see Appendix~\ref{app:amortization}).

\subsection{Downstream Question Answering}
Once $T_{\text{C}}$ is produced, downstream question answering
proceeds over the canonical table without further modification.
Given a query $q$, the reasoning model $M$ generates a
chain-of-thought reasoning trace that decomposes the question into
sub-steps, then translates it into a SQL query executed against
$T_{\text{C}}$ to produce the final answer $\hat{a}$. Because
$T_{\text{C}}$ already conforms to $\mathcal{S}_{\pi}$, this
CoT$\rightarrow$SQL pipeline operates over a consistent representation
regardless of the original table's structural complexity.


\section{\textsc{QuIeTT} Challenge Set}
\label{sec:challenge_set}


To evaluate generalization under increased structural 
complexity, we construct a challenge set from 
WikiTQ~\cite{pasupat-liang-2015-compositional} and 
NQ-Table~\cite{kwiatkowski-etal-2019-natural}, two 
datasets whose structural issues, heterogeneous formats, 
implicit relations, and irregular schemas are latent and 
unlabeled, making them the most suitable sources for 
stress-testing table-intrinsic representation. 

\paragraph{Dataset Creation.}
For each table, we generate structurally challenging 
question-answer pairs directly from the raw table, targeting 
structural patterns that hinder reliable answering without 
transformation, following prior LLM-based table QA 
benchmarks~\cite{choudhury-etal-2025-tabard,
tian2025tabulargsmunderstandinglimitationsllms,
choudhury2026betterclausediscrepancybenchmark}. We acknowledge 
that generating questions from raw tables may not surface all 
structural complexity compared to canonical representations, 
but ensures question generation is independent of 
\textsc{QuIeTT}'s transformation. To ensure evaluation 
neutrality, we apply two controls. First, all answers must be 
derivable exclusively from the source raw table. Second, all 
question-answer pairs are manually verified by three 
independent NLP annotators against the raw table, without 
access to the canonical representation or model outputs. Because 
all methods share the same fixed question set and 
\textsc{QuIeTT} has no access to these questions during 
transformation, any representational alignment introduced during 
question generation cannot advantage \textsc{QuIeTT} at 
evaluation time.
\paragraph{Dataset Verification.}
Three independent NLP annotators validate each question-answer 
pair against the raw source table. 
Table~\ref{tab:inter_rater_agreement} reports inter-annotator 
agreement, which is consistently high across all pairs, 
supporting the reliability of the challenge set.

\begin{table}[t]
\vspace{-0.5em}
\small
\centering
\setlength{\aboverulesep}{0.1pt}
\setlength{\belowrulesep}{0.1pt}
\setlength\tabcolsep{3.75pt}
\begin{tabular}{lcc}
\toprule
\textbf{Expert Pair} & \textbf{Cohen's Kappa} & \textbf{Jaccard Coeff.} \\
\midrule
$\alpha_{1,2}$ & 92.9 & 96.9 \\
$\alpha_{2,3}$ & 92.0 & 96.6 \\
$\alpha_{1,3}$ & 94.9 & 97.7 \\
$\alpha_{1,2,3}$ (Fleiss' $\kappa$) & 0.9328 & 0.9562 \\
\bottomrule
\end{tabular}
\vspace{-0.75em}
\caption{\small Inter-rater agreement for challenge set annotation.}
\label{tab:inter_rater_agreement}
\vspace{-1.0em}
\end{table}

\subsection{Dataset Statistics.}
Table~\ref{tab:challenge_overview}
details scale and question type distribution.

\begin{table}[tb]
\vspace{-0.5em}
\small
\centering
\begin{tabular}{lcl}
\toprule
\textbf{Statistic} & \textbf{Values} & \textbf{Table Dimension}\\
\midrule
Total tables & 529 & Mean: $(20,\,11)$ \\
Total questions & 2{,}500 & Median: $(8,\,8)$ \\
Questions per table & 4 / 5 / 7 & Min: $(1,\,1)$ \\
(mean / median / max) & & Max: $(380,\,76)$ \\
\bottomrule
\end{tabular}
\vspace{-0.75em}
\caption{\small \textbf{Challenge set composition and table structure statistics.}
The set comprises 2,500 questions over 529 tables.
Tuple-valued entries in the Table Dimension column report (rows,~columns); the
wide size range reflects the structural diversity the challenge set targets.}
\label{tab:challenge_overview}
\vspace{-1.0em}
\end{table}

We evaluate \textsc{QuIeTT} across four benchmarks, five model
families, and two reasoning paradigms, holding downstream QA
models and prompting strategies fixed across all conditions.

\section{Experiments}
\subsection{Benchmark Datasets.}
We conduct experiments on four widely used table QA benchmarks
covering diverse table structures and reasoning requirements.
WikiTQ ~\cite{pasupat-liang-2015-compositional} consists
of semi-structured Wikipedia tables paired with natural language
questions, emphasizing lookup and aggregation.
NQ-Table ~\cite{kwiatkowski-etal-2019-natural} is derived
from Natural Questions and contains noisier web tables with
implicit schemas.
SequentialQA ~\cite{iyyer-etal-2017-search} requires
multi-turn reasoning where later questions depend on earlier
answers.
HiTab ~\cite{cheng-etal-2022-hitab} focuses on
hierarchical and numerical reasoning over complex table structures.
We additionally evaluate on the challenge set constructed from
WikiTQ and NQ-Table (Section~\ref{sec:challenge_set}) to assess
robustness on structurally harder cases. WikiTQ and NQ-Table are selected for the challenge set because 
their structural issues, heterogeneous formats, implicit 
relations, and irregular schemas are latent and unlabeled, 
making them the most suitable sources for stress testing 
table-intrinsic representation quality. HiTab's difficulty is 
well characterized by its explicit hierarchical structure and 
SequentialQA's by conversational context, so neither isolates 
the schema-level variation the challenge set targets.

\subsection{Evaluation Metrics.}
We report both token-level F1 and accuracy across
all datasets. Before scoring, we apply lightweight surface
normalization to both predictions and gold answers such as date strings are
converted to a canonical format (\textit{e.g.}, \texttt{``Jan 5,
2018''} $\rightarrow$ \texttt{``2018-01-05''}), non-semantic
characters such as \texttt{@}, \texttt{;}, and \texttt{|} are removed,
and Roman numerals are mapped to integers (\textit{e.g.},
\texttt{``III''} $\rightarrow$ \texttt{``3''}). Token-level F1 is then
computed over the normalized tokens, and accuracy is computed as exact
match after the same normalization. We report both metrics because most
table QA benchmarks and prior methods report accuracy as their primary
metric, while F1 additionally captures graded recovery for list-valued
and compositional answers where a prediction may be partially correct.
All methods are evaluated using the same normalization and scoring
scripts.

\subsection{Models and Baselines.}
We evaluate across five large language models:
Gemini-2.0-flash and Gemini-2.5-flash~\cite{gemini2023},
DeepSeek-V3.1~\cite{deepseek2024}, Qwen3-80B,
and GPT-OSS-120B\footnote{Qwen3-80B and GPT-OSS-120B are 
open-weight models accessed via Vertex AI 
Model-as-a-Service. GPT-OSS-120B is a 120B-parameter 
model from OpenAI; Qwen3-80B is an 80B-parameter model.}, spanning both proprietary and open-weight
settings. \textbf{Baselines} include prompting-based methods
(CoT, CoT+SQL, E5~\cite{zhang2024e5},
EEDP~\cite{srivastava2024financialqa}), normalization and
restructuring methods (NormTab~\cite{nahid2025normtab},
TableSQLify~\cite{nahid2024tablesqlify},
TableReasoner~\cite{xiong2025tablereasoneradvancingtablereasoning},
GraphOTTER~\cite{li2024graphotterevolvingllmbasedgraph}),
an execution-based system (Binder~\cite{Binder}),
and multi-step agentic methods
(Chain-of-Table~\cite{wang2024chainoftable},
Program-of-Thoughts). All baselines are implemented following their original papers and 
publicly released code, using default hyperparameters and prompting 
strategies without additional tuning.

\begin{table*}[!t]
\centering
\scriptsize
\setlength{\aboverulesep}{0.1pt}
\setlength{\belowrulesep}{0.1pt}
\setlength\tabcolsep{2.5pt}
\begin{tabular}{c l cc cc cc cc cc cc cc cc cc}
\toprule
\rowcolor{quiettgrey}
& & \multicolumn{8}{c}{\textbf{Gemini 2.5}} & \multicolumn{8}{c}{\textbf{GPT-OSS}} \\
\cmidrule(lr){3-10}\cmidrule(lr){11-18}
\multirow{2}{*}{\bf Category} & \multirow{2}{*}{\bf Method}
& \multicolumn{2}{c}{\textbf{WikiTQ}}
& \multicolumn{2}{c}{\textbf{NQ-Table}}
& \multicolumn{2}{c}{\textbf{SequentialQA}}
& \multicolumn{2}{c}{\textbf{HiTab}}
& \multicolumn{2}{c}{\textbf{WikiTQ}}
& \multicolumn{2}{c}{\textbf{NQ-Table}}
& \multicolumn{2}{c}{\textbf{SequentialQA}}
& \multicolumn{2}{c}{\textbf{HiTab}} \\
\cmidrule(lr){3-4}\cmidrule(lr){5-6}\cmidrule(lr){7-8}\cmidrule(lr){9-10}
\cmidrule(lr){11-12}\cmidrule(lr){13-14}\cmidrule(lr){15-16}\cmidrule(lr){17-18}
& & \bf F1 & \bf EM & \bf F1 & \bf EM & \bf F1 & \bf EM & \bf F1 & \bf EM
& \bf F1 & \bf EM & \bf F1 & \bf EM & \bf F1 & \bf EM & \bf F1 & \bf EM \\
\midrule
\multirow{8}{*}{\textbf{Direct}}
& CoT           & 74.31 & 72.85 & 71.22 & 66.36 & 61.98 & 57.21 & 75.21 & 71.00 & 69.40 & 66.56 & 66.60 & 63.54 & 60.26 & 58.29 & 53.90 & 50.80 \\
& CoT + SQL     & 54.66 & 52.49 & 36.64 & 30.96 & 58.23 & 53.37 & 69.01 & 67.00 & 43.00 & 42.29 & 47.90 & 33.95 & 55.20 & 49.60 & 31.90 & 30.70 \\
& NormTab       & 58.70 & 56.19 & 53.00 & 51.05 & 50.00 & 49.92 & 56.81 & 54.60 & 65.91 & 62.61 & 50.30 & 48.34 & 48.60 & 43.10 & 54.67 & 52.00 \\
& E5            & 74.35 & 70.30 & 55.90 & 52.92 & 31.64 & 29.81 & 78.89 & 75.10 & 56.90 & 52.15 & 54.00 & 51.76 & 43.80 & 40.46 & 77.47 & 75.00 \\
& EEDP          & 65.00 & 51.10 & 72.79 & 70.17 & 60.30 & 57.13 & 77.60 & 73.00 & 60.10 & 59.54 & 71.80 & 58.28 & 54.90 & 50.31 & 55.48 & 52.40 \\
& Binder        & 51.89 & 49.69 & 67.24 & 59.59 & 22.49 & 19.84 & 59.03 & 56.30 & 61.50 & 58.47 & 66.50 & 64.30 & 21.50 & 17.23 & 58.55 & 55.03 \\
& TableSQLify   & 53.02 & 49.50 & 34.78 & 27.77 & 26.78 & 24.44 & 69.92 & 67.90 & 73.90 & 68.98 & 46.99 & 41.52 & 26.73 & 23.87 & 53.00 & 40.41 \\
& TableReasoner & 54.32 & 50.05 & 64.71 & 57.58 & 42.51 & 39.15 & 61.68 & 59.54 & 69.79 & 64.64 & 62.70 & 59.45 & 66.22 & 63.37 & 56.81 & 52.78 \\
\cmidrule(lr){2-18}
\multirow{4}{*}{\textbf{Agentic}}
& Chain-of-Table & 74.30 & 73.20 & 69.56 & 67.45 & 22.27 & 20.93 & 70.12 & 65.80 & 62.11 & 57.53 & 60.21 & 53.47 & 23.20 & 19.14 & 65.60 & 53.80 \\
& POT            & 60.21 & 58.39 & 67.70 & 65.69 & 63.52 & 54.58 & 69.12 & 65.40 & 69.90 & 68.16 & 57.60 & 50.08 & 44.50 & 41.84 & 43.50 & 38.23 \\
& GraphOTTER     & 49.29 & 42.75 & 61.79 & 56.09 & 42.76 & 43.98 & 81.05 & 79.71 & 57.50 & 52.50 & 55.88 & 48.47 & 44.26 & 42.26 & 78.60 & 74.61 \\
\cmidrule(lr){2-18}
\rowcolor{quiettgreen}
& \textbf{\textsc{QuIeTT}}
& \textbf{79.80} & \textbf{78.56}
& \textbf{80.10} & \textbf{76.19}
& \textbf{72.32} & \textbf{66.69}
& \textbf{84.41} & \textbf{83.10}
& \textbf{76.21} & \textbf{73.23}
& \textbf{72.75} & \textbf{68.50}
& \textbf{73.00} & \textbf{64.58}
& \textbf{79.50} & \textbf{78.70} \\
\specialrule{1.2pt}{0pt}{0pt}
\rowcolor{quiettgrey}
& & \multicolumn{8}{c}{\textbf{DeepSeek V3.1}} & \multicolumn{8}{c}{\textbf{Qwen 3}} \\
\cmidrule(lr){3-10}\cmidrule(lr){11-18}
& & \multicolumn{2}{c}{\textbf{WikiTQ}}
& \multicolumn{2}{c}{\textbf{NQ-Table}}
& \multicolumn{2}{c}{\textbf{SequentialQA}}
& \multicolumn{2}{c}{\textbf{HiTab}}
& \multicolumn{2}{c}{\textbf{WikiTQ}}
& \multicolumn{2}{c}{\textbf{NQ-Table}}
& \multicolumn{2}{c}{\textbf{SequentialQA}}
& \multicolumn{2}{c}{\textbf{HiTab}} \\
\cmidrule(lr){3-4}\cmidrule(lr){5-6}\cmidrule(lr){7-8}\cmidrule(lr){9-10}
\cmidrule(lr){11-12}\cmidrule(lr){13-14}\cmidrule(lr){15-16}\cmidrule(lr){17-18}
& & \bf F1 & \bf EM & \bf F1 & \bf EM & \bf F1 & \bf EM & \bf F1 & \bf EM
& \bf F1 & \bf EM & \bf F1 & \bf EM & \bf F1 & \bf EM & \bf F1 & \bf EM \\
\midrule
\multirow{8}{*}{\textbf{Direct}}
& CoT           & 60.92 & 59.62 & 63.60 & 59.12 & 59.50 & 50.36 & 71.13 & 68.00 & 58.95 & 57.55 & 61.34 & 59.91 & 57.12 & 54.03 & 72.97 & 69.00 \\
& CoT + SQL     & 51.21 & 47.39 & 34.21 & 29.28 & 56.30 & 55.12 & 65.45 & 61.00 & 51.55 & 48.95 & 34.54 & 33.36 & 59.64 & 58.11 & 56.23 & 51.00 \\
& NormTab       & 53.19 & 51.33 & 55.40 & 52.34 & 45.57 & 43.24 & 43.35 & 42.27 & 51.12 & 47.72 & 46.80 & 41.92 & 48.68 & 46.69 & 43.30 & 42.78 \\
& E5            & 73.00 & 68.20 & 54.55 & 49.45 & 42.75 & 35.11 & 80.50 & 76.80 & 62.30 & 57.30 & 53.16 & 50.23 & 42.69 & 33.97 & 72.21 & 71.20 \\
& EEDP          & 62.20 & 52.40 & 72.29 & 69.42 & 60.25 & 52.79 & 74.18 & 69.60 & 62.40 & 53.70 & 62.21 & 61.31 & 55.90 & 53.18 & 71.27 & 67.70 \\
& Binder        & 50.62 & 47.22 & 67.44 & 61.20 & 20.19 & 16.63 & 58.66 & 55.90 & 61.68 & 58.48 & 61.56 & 59.38 & 20.54 & 20.34 & 56.29 & 53.50 \\
& TableSQLify   & 73.00 & 71.45 & 56.94 & 48.03 & 30.01 & 28.58 & 67.62 & 63.40 & 48.40 & 45.45 & 51.34 & 47.67 & 29.18 & 27.27 & 62.30 & 60.40 \\
& TableReasoner & 47.63 & 41.38 & 33.33 & 32.36 & 56.89 & 50.36 & 48.63 & 46.42 & 62.49 & \textbf{61.53} & 63.77 & 60.11 & 60.59 & 58.99 & 52.95 & 49.18 \\
\cmidrule(lr){2-18}
\multirow{4}{*}{\textbf{Agentic}}
& Chain-of-Table & 63.67 & 61.29 & 70.10 & 61.23 & 44.53 & 41.66 & 75.55 & 70.80 & 61.39 & 59.17 & 64.33 & 61.20 & 44.28 & 38.88 & 76.18 & 71.80 \\
& POT            & 48.39 & 45.72 & 65.33 & 58.68 & 59.79 & 56.47 & 56.55 & 53.30 & 46.56 & 44.33 & 62.10 & 59.68 & 59.28 & 53.12 & 32.30 & 30.70 \\
& GraphOTTER     & 34.80 & 30.30 & 35.29 & 32.63 & 44.05 & 39.73 & 61.30 & 58.82 & 50.84 & 46.94 & 55.80 & 51.41 & 42.93 & 40.53 & 77.05 & 73.63 \\
\cmidrule(lr){2-18}
\rowcolor{quiettgreen}
& \textbf{\textsc{QuIeTT}}
& \textbf{74.56} & \textbf{72.21}
& \textbf{79.20} & \textbf{72.29}
& \textbf{66.15} & \textbf{59.57}
& \textbf{81.30} & \textbf{79.40}
& \textbf{64.20} & 60.96
& \textbf{72.60} & \textbf{65.23}
& \textbf{62.54} & \textbf{58.39}
& \textbf{81.22} & \textbf{81.00} \\
\bottomrule
\end{tabular}
\vspace{-0.9em}
\caption{\small \textbf{Main results.} \textsc{QuIeTT} 
is compared against direct prompting methods and 
agentic methods across four 
benchmarks and four model families. Top half: Gemini~2.5 and 
GPT-OSS; bottom half: DeepSeek~V3.1 and Qwen~3. Bold indicates 
best result per column. Gemini~2.0 results are in 
Table~\ref{tab:gemini20}.}
\label{tab:main_results}
\vspace{-1.5em}
\end{table*}

\section{Results and Analysis}
We structure our analysis around four questions:
\textbf{(1)}~Does \textsc{QuIeTT} improve consistently across model families and scales?
\textbf{(2)}~Do gains generalize across diverse table structures and reasoning settings?
\textbf{(3)}~Is transformation quality a stronger driver of performance or QA model capacity?
\textbf{(4)}~How does \textsc{QuIeTT} perform on structurally challenging unseen queries?
\paragraph{Does \textsc{QuIeTT} generalize across model families?}
\textsc{QuIeTT} consistently improves over the strongest baseline for
each model across both F1 and accuracy, with F1 gains of approximately
4-8 points and corresponding accuracy gains of 3-6 points across
datasets. These improvements hold for both proprietary models
(Gemini~2.0, Gemini~2.5) and open-weight models (DeepSeek-V3.1,
Qwen3-80B, GPT-OSS-120B), indicating that the benefits of improved
table representation generalize across model families and scales. 
\paragraph{Does \textsc{QuIeTT} generalize across table structures?}
\textsc{QuIeTT} generalizes across diverse table types and reasoning
settings. On WikiTQ and NQ-Table, which contain semi-structured web
tables, \textsc{QuIeTT} reaches F1 of
79.80 and 80.10 with accuracy of 78.56 and 76.19 respectively with
Gemini~2.5. On SequentialQA, where questions are inter-dependent
across turns, \textsc{QuIeTT} achieves up to 72.32 F1 and 66.69
accuracy, showing that preprocessing benefits extend beyond
single-turn answering. On HiTab, which emphasizes hierarchical
numerical reasoning, \textsc{QuIeTT} achieves peak F1 of 84.41 and
accuracy of 83.10, the highest across all methods and models.
Together, these results suggest that table-intrinsic structural issues
are a consistent bottleneck across table types, and that resolving
them before reasoning is broadly beneficial.

\vspace{-0.20em} 
\paragraph{Does transformation quality matter more or model capacity?}
Table~\ref{tab:transformation_acl} shows that transformation quality
is a stronger driver of downstream performance than QA model capacity.
On WikiTQ, tables transformed with Gemini~2.5 achieve 65.12 F1 and
62.23 accuracy with Mistral, compared to 63.33 F1 and 59.35 accuracy
when Qwen~3 performs the transformation with the same QA model.
Similar trends hold on NQ-Table (63.28 \& 58.12 vs.\ 62.56 \& 60.23)
and SequentialQA (47.24 \& 44.84 vs.\ 42.81 \& 41.12). The pipeline is
also not critically dependent on large models for issue detection,
replacing the issue probing model with LLaMA-14B Scout
(Table~\ref{tab:ablation_slm}) achieves 68.34 F1 and 68.14 accuracy
on WikiTQ and 70.11 F1 and 69.34 accuracy on NQ-Table with
Gemini~2.5, within 6-10 scores across both metrics, suggesting that
structural deficiencies are detectable even by smaller models.
\paragraph{How are gains on structurally challenging unseen queries?}
Table~\ref{tab:challenge_results} reports performance on the challenge
set described in Section~\ref{sec:challenge_set}. \textsc{QuIeTT}
achieves 74.41 F1 and 71.37 accuracy on WikiTQ and 77.16 F1 and
76.50 accuracy on NQ-Table with Gemini~2.5, outperforming all
baselines by 5-8 F1 and 4-7 accuracy points. These gains are larger
than those on the standard benchmarks, consistent with the challenge
set targeting structurally diverse cases where table-intrinsic
representation issues are most acute. The rationale for selecting
WikiTQ and NQ-Table for the challenge set is discussed in
Section~\ref{sec:challenge_set}.

\begin{table}[!htb]
\small
\setlength{\aboverulesep}{0.1pt}
\setlength{\belowrulesep}{0.1pt}
\setlength\tabcolsep{3pt}
\centering
\begin{tabular}{l l c c c c}
\toprule
\rowcolor{quiettgrey}
\textbf{Dataset} & \textbf{Method}
& \multicolumn{2}{c}{\textbf{Gemini 2.5}}
& \multicolumn{2}{c}{\textbf{GPT-OSS}} \\
\cmidrule(lr){3-4} \cmidrule(lr){5-6}
& & \textbf{F1} & \textbf{EM}
& \textbf{F1} & \textbf{EM} \\
\midrule
\multirow{5}{*}{WikiTQ}
& CoT      & 67.11 & 65.50 & 63.58 & 60.36 \\
& NormTab  & 62.28 & 61.90 & 61.71 & 57.23 \\
& EEDP     & 67.60 & 66.34 & 65.12 & 61.86 \\
& Binder   & 65.56 & 63.39 & 65.12 & 64.48 \\
\cmidrule(lr){2-6}
\rowcolor{quiettgreen}
& \textbf{\textsc{QuIeTT}} & \textbf{74.41} & \textbf{71.37} & \textbf{70.06} & \textbf{68.59} \\
\midrule
\multirow{5}{*}{NQ-Table}
& CoT      & 65.63 & 63.35 & 61.38 & 58.49 \\
& NormTab  & 66.84 & 66.12 & 65.68 & 64.23 \\
& EEDP     & 71.94 & 68.50 & 67.18 & 63.87\\
& Binder   & 72.28 & 67.22 & 68.33 & 67.30 \\
\cmidrule(lr){2-6}
\rowcolor{quiettgreen}
& \textbf{\textsc{QuIeTT}} & \textbf{77.16} & \textbf{76.50} & \textbf{71.19} & \textbf{70.66} \\
\bottomrule
\end{tabular}
\vspace{-0.75em}
\caption{\small \textbf{Challenge set results}}
\label{tab:challenge_results}
\vspace{-1.0em}
\end{table}

\begin{figure}[!htb]
\centering
\includegraphics[width=0.95\columnwidth]{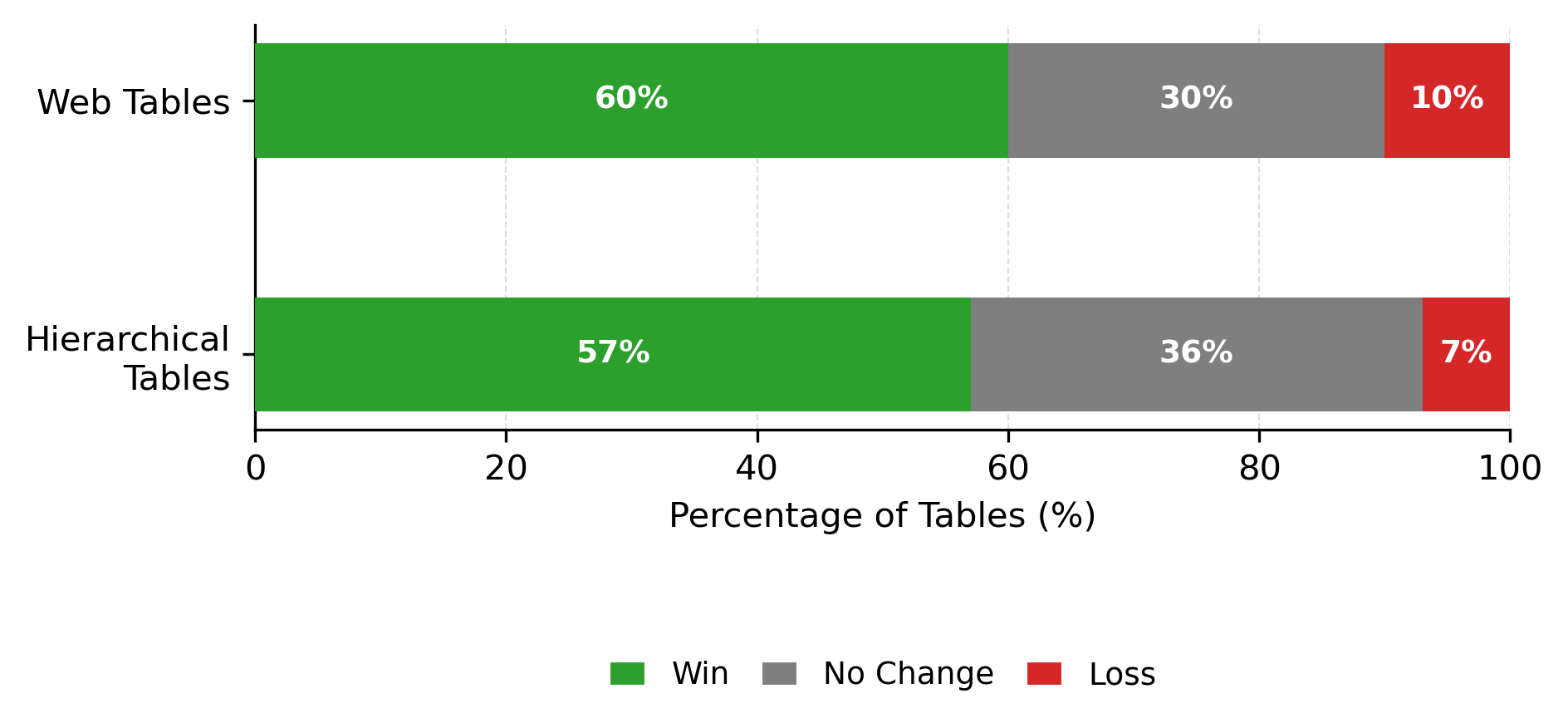}
\caption{\small \textbf{Table-level performance outcomes.} Win, No Change, and Loss indicate whether \textsc{QuIeTT} improves, leaves unchanged, or degrades per-table QA performance relative to the raw table baseline.}
\label{fig:table_level_changes}
\end{figure}

\paragraph{How often does \textsc{QuIeTT} improve vs.\ degrade per table?}
Figure~\ref{fig:table_level_changes} shows per-table
outcomes after applying \textsc{QuIeTT}. On web tables,
60\% improve, 31\% show no change, and 9\% degrade. On
HiTab, 57\% improve, 36\% show no change, and 7\% degrade.
The higher proportion of unchanged cases on HiTab reflects
that hierarchical tables often already encode relational
structure explicitly. Provenance columns are introduced
in only 13\% of web tables and 9\% of HiTab tables,
confirming that most canonical representations do not
rely on auxiliary columns for information completeness.

\begin{figure}[!htb]
\centering
\includegraphics[width=\columnwidth]{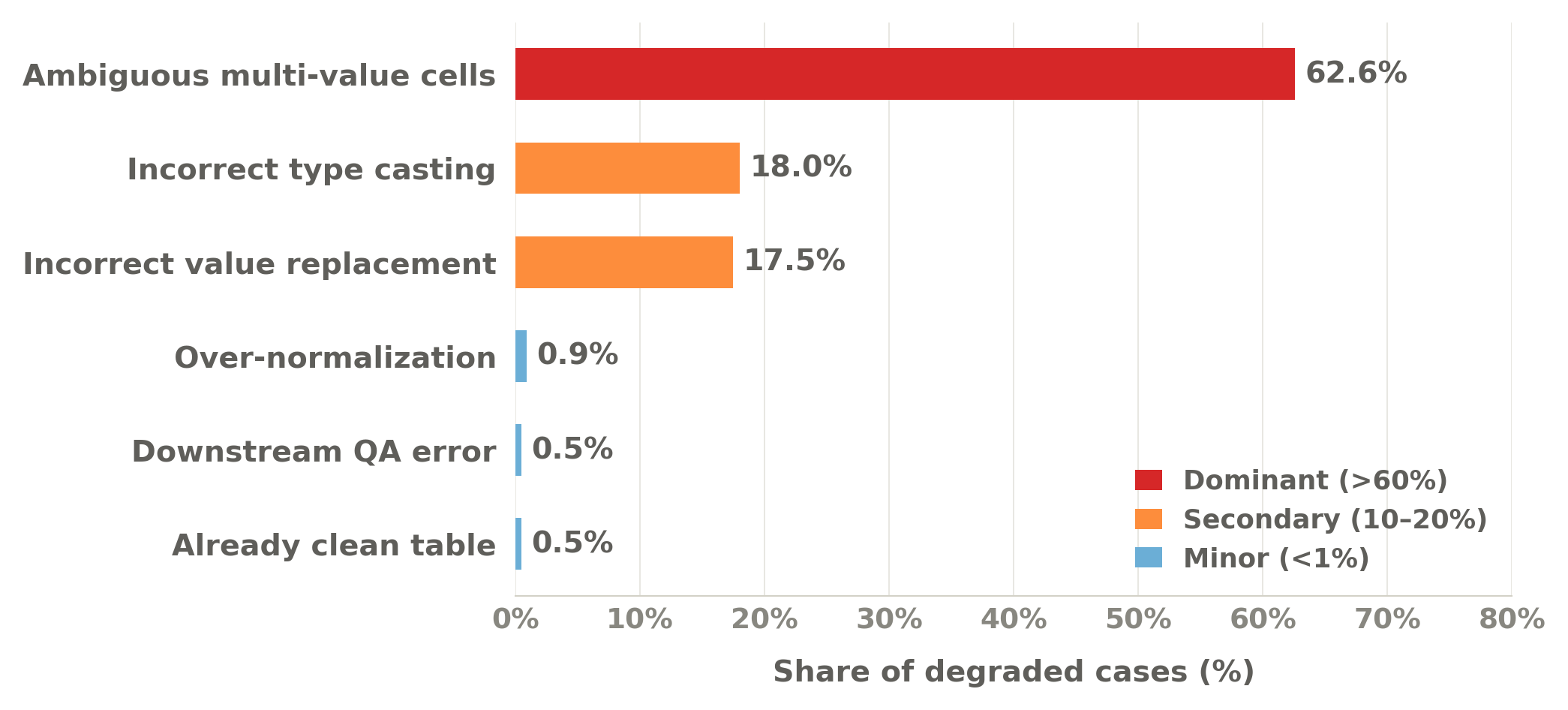}
\caption{\small \textbf{Error Analysis for degraded tables.}}
\label{fig:error_taxonomy}
\end{figure}

\paragraph{When does transformation hurt?}
To characterize the 7-9\% of tables where performance 
degrades (Figure~\ref{fig:table_level_changes}), we analyze all degraded cases across both 
web and hierarchical tables (Figure~\ref{fig:error_taxonomy}). 
The dominant failure mode (62.6\%) is ambiguous
multi-value cells or packed cells that are difficult
to split cleanly, causing answer spans to become
partial or harder to recover. The remaining failures
arise from surface-form changes like type casting (18.0\%)
alters matching behavior, and value replacement
(17.5\%) occasionally normalizes a value that was
useful for answer extraction. Over-normalization,
downstream QA errors on valid tables, and unnecessary
transformation of already-clean tables together account
for under 2\% of degraded cases. These failure modes are localized and interpretable, 
reflecting inherent ambiguities in specific tables 
rather than a systematic weakness of the 
transform-first approach. The 7-9\% degradation also provides a natural signal for identifying 
tables that may benefit from complementary 
query-conditioned methods.

\subsection{Ablation studies.}
We conduct three targeted ablations to isolate the
contribution of individual pipeline components.

\paragraph{Removing issue probing.}
Skipping the issue probing stage entirely
(Table~\ref{tab:ablation_no_issue}) causes the largest performance
drops: from 79.80 to 53.22 F1 and 78.56 to 52.06 accuracy on WikiTQ,
and from 80.10 to 56.80 F1 and 76.19 to 54.37 accuracy on NQ-Table
with Gemini~2.5, decreases exceeding 25 F1 and accuracy points.
Similar drops are observed with GPT-OSS, falling to 51.40 F1 and
51.15 accuracy on WikiTQ and 54.90 F1 and 51.82 accuracy on NQ-Table.
Without issue detection, the transformation plan has no structured
basis for deciding what to normalize or derive, and the pipeline
reduces to undirected table restructuring.

\begin{table}[t]
\centering
\small
\setlength{\tabcolsep}{4pt}
\renewcommand{\arraystretch}{1.1}
\begin{tabular}{lcccc}
\toprule
\multirow{2}{*}{\textbf{Dataset}} &
\multicolumn{2}{c}{\textbf{Gemini 2.5}} &
\multicolumn{2}{c}{\textbf{GPT-OSS}} \\
\cmidrule(lr){2-3} \cmidrule(lr){4-5}
 & \textbf{F1} & \textbf{EM} & \textbf{F1} & \textbf{EM} \\
\midrule
WikiTQ   & 53.22 & 52.06 & 51.40 & 51.15 \\
NQ-Table & 56.80 & 54.37 & 54.90 & 51.82 \\
\bottomrule
\end{tabular}
\caption{\small \textsc{QuIeTT} performance without issue probing.}
\label{tab:ablation_no_issue}
\end{table}

\paragraph{Merging planning and execution.}
Collapsing plan generation and code execution into a
single-agent pipeline (Table~\ref{tab:ablation_single_agent})
degrades performance consistently across both models and
metrics. On WikiTQ, performance drops from 79.80 to 60.57 F1
and 78.56 to 58.26 accuracy with Gemini~2.5, and from 76.21
to 58.33 F1 and 73.23 to 57.29 accuracy with GPT-OSS.
NQ-Table follows the same pattern, falling from 80.10 to 63.20
F1 and 76.19 to 60.56 accuracy with Gemini~2.5, and from 72.75
to 61.08 F1 and 68.50 to 58.66 accuracy with GPT-OSS. This
confirms that separating symbolic planning from code-level
execution provides meaningful structure. The plan constrains
\emph{what} is transformed, and the executor enforces
\emph{how}.

\begin{table}[t]
\centering
\small
\setlength{\tabcolsep}{4pt}
\renewcommand{\arraystretch}{1.1}
\begin{tabular}{lcccc}
\toprule
\multirow{2}{*}{\textbf{Dataset}} &
\multicolumn{2}{c}{\textbf{Gemini 2.5}} &
\multicolumn{2}{c}{\textbf{GPT-OSS}} \\
\cmidrule(lr){2-3} \cmidrule(lr){4-5}
 & \textbf{F1} & \textbf{EM} & \textbf{F1} & \textbf{EM} \\
\midrule
WikiTQ   & 60.57 & 58.26 & 58.33 & 57.29 \\
NQ-Table & 63.20 & 60.56 & 61.08 & 58.66 \\
\bottomrule
\end{tabular}
\caption{\small \textsc{QuIeTT} performance when planning and
execution are merged into a single-agent transformation pipeline.}
\label{tab:ablation_single_agent}
\end{table}

\paragraph{Performance under constrained inputs.}
Table~\ref{tab:main_results_simple} reports performance when QA
models receive only the schema, data types, and two representative
rows ($n=2$) of the transformed table, without access to the full
table, raw tables, or execution traces. Performance decreases
relative to full-context settings, as expected, but
\textsc{QuIeTT} retains a substantial fraction of its gains
across all datasets and models, with HiTab showing the strongest
retention (62.30 F1 and 59.60 accuracy with Gemini~2.5) while WikiTQ and NQ-Table drop more sharply, suggesting that web tables 
benefit more from full table access. Nevertheless, the canonical 
representation remains more useful than raw tables under the 
same constraints, indicating that structural normalization 
provides value beyond simply exposing more content.

\begin{table}[t]
\centering
\scriptsize
\setlength{\aboverulesep}{0.1pt}
\setlength{\belowrulesep}{0.1pt}
\setlength{\tabcolsep}{2pt}
\renewcommand{\arraystretch}{1.4}
\resizebox{\columnwidth}{!}{%
\begin{tabular}{l cc cc cc cc cc}
\toprule
\textbf{Dataset}
& \multicolumn{2}{c}{\textbf{Gemini 2.0}}
& \multicolumn{2}{c}{\textbf{Gemini 2.5}}
& \multicolumn{2}{c}{\textbf{DeepSeek V3.1}}
& \multicolumn{2}{c}{\textbf{Qwen 3}}
& \multicolumn{2}{c}{\textbf{GPT-OSS}} \\
\cmidrule(lr){2-3} \cmidrule(lr){4-5} \cmidrule(lr){6-7}
\cmidrule(lr){8-9} \cmidrule(lr){10-11}
& \textbf{F1} & \textbf{EM}
& \textbf{F1} & \textbf{EM}
& \textbf{F1} & \textbf{EM}
& \textbf{F1} & \textbf{EM}
& \textbf{F1} & \textbf{EM} \\
\midrule
WikiTQ       & 37.61 & 36.34 & 38.92 & 35.12 & 27.65 & 26.78 & 27.19 & 23.50 & 28.21 & 24.59 \\
NQ-Table     & 35.42 & 33.32 & 32.49 & 32.03 & 25.38 & 24.23 & 35.31 & 31.21 & 32.44 & 30.46 \\
SequentialQA & 40.71 & 38.90 & 42.97 & 36.32 & 38.98 & 35.33 & 36.67 & 31.12 & 41.98 & 38.48 \\
HiTab        & 59.60 & 56.80 & 62.30 & 59.60 & 58.12 & 58.30 & 46.27 & 45.30 & 53.60 & 52.50 \\
\bottomrule
\end{tabular}}
\caption{\small\textsc{QuIeTT} with limited table representations (n=2).}
\label{tab:main_results_simple}
\end{table}

\section{Related Work}
\label{sec:related_work}

Table QA methods differ primarily in \textit{when} 
and \textit{how} they address structural issues in 
raw tables. We organize prior work along three axes: 
lightweight normalization, programmatic restructuring, 
and transformation coupled with reasoning, and 
clarify where \textsc{QuIeTT} sits relative to each.

\paragraph{Lightweight normalization for table QA.}
Some methods apply one-time normalization to reduce surface-form
variability while preserving the original structure.
NormTab~\cite{nahid2025normtab} standardizes values such as
dates and numbers and applies limited structural fixes, but
leaves hierarchical structure, implicit relations, and
heterogeneous layouts to downstream reasoning. Related approaches
similarly assume mostly fixed schemas and rely on query-time
reasoning to resolve remaining
ambiguity~\cite{pasupat-liang-2015-compositional,
nahid2024tablesqlify}. Unlike these methods, \textsc{QuIeTT}
resolves structural issues proactively, before any query is
observed, and enforces a single reusable canonical representation.

\paragraph{Programmatic table restructuring.}
A second line of work performs restructuring via explicit
transformation programs. Auto-Tables~\cite{10.14778/3611479.3611534}
synthesizes multi-step pipelines from a fixed operator vocabulary
but is limited to predefined transformations and known target
schemas. DataMorpher~\cite{11112951} and
RelationalCoder~\cite{dong-etal-2025-relationalcoder} generate
Python or SQL programs for ETL or schema matching, but remain
coupled to user-specified target schemas and do not enforce
reusable, query-independent representations. \textsc{QuIeTT}
differs in that no target schema is predefined, and transformation
is evaluated by downstream QA improvement across arbitrary
queries and models.

\paragraph{Transformation coupled with reasoning.}
Other approaches integrate transformation directly into reasoning.
TabFormer~\cite{dong2025rrdt} jointly applies restructuring and
SQL answering within a single prompted process, making
transformations dependent on the query set.
Chain-of-Table~\cite{wang2024chainoftable} and
TableReasoner~\cite{xiong-etal-2025-teleai} adopt query-driven
schema selection or iterative reasoning, improving efficiency but
coupling table structure to query-time decisions. In contrast,
\textsc{QuIeTT} treats transformation as a query-independent
preprocessing step that produces a single canonical representation
reused across queries, tasks, and models, cleanly separating
representation quality from downstream reasoning. A detailed
discussion of the limitations of query-conditioned transformation is provided in Appendix~\ref{app:query_conditioned}.


\vspace{-0.25em} 
\section{Conclusion}
\label{sec:conclusion}
\textsc{QuIeTT} converts raw, semi-structured tables into a single 
SQL-ready canonical representation before any evaluation query is 
observed, decoupling table pre-processing from downstream reasoning. 
Experiments across four benchmarks and five model families show 
consistent gains in F1 and accuracy under identical prompting 
conditions, including with smaller models and constrained inputs, 
indicating that structural normalization of the table itself, 
independent of the query, is a reliable and underexplored source 
of improvement in table QA. These results suggest that the field has focused heavily on 
query-adaptive reasoning while leaving systematic table 
preparation largely unaddressed. A single reusable canonical 
representation, constructed without query access, can serve 
as a stable foundation for diverse reasoning strategies, 
models, and tasks. Future work should extend this direction 
to large tables, multimodal formats, and multi-table 
reasoning settings.

\section*{Limitations}
\label{sec:limitations}
\textsc{QuIeTT} has five limitations. First, the 
pipeline targets tables processable within current context windows; 
very large tables remain challenging due to input length constraints 
and the difficulty of producing compact canonical representations. 
Second, \textsc{QuIeTT} operates on structured tabular data only 
and does not handle tables interleaving free-form text or images. 
Third, while the canonical representation enables effective QA with 
smaller downstream models, the transformation stage itself requires 
sufficiently capable models for issue probing and plan generation. 
Fourth, a single fixed transformation cannot anticipate every future 
query; cases requiring latent relations not surfaced during probing 
may still benefit from query-conditioned methods such as 
Chain-of-Table~\cite{wang2024chainoftable}, and \textsc{QuIeTT} 
works best alongside rather than instead of such approaches. 
Finally, ambiguous packed cells remain the dominant failure mode 
(62.6\% of degraded cases, Figure~\ref{fig:error_taxonomy}); more 
principled alternatives such as child-table 
relationalization~\cite{10.14778/3611479.3611534,dong-etal-2025-relationalcoder} 
would handle these more robustly but require relaxing the 
single-table reuse constraint~\ref{sec:problem_setup}~(c).

\section*{Ethics Statement}
\label{sec:Ethics}

This work studies table transformation and reasoning over publicly available benchmark datasets, including WikiTQ, HiTab, NQ-Table, and SequentialQA. All datasets consist of tabular data derived from publicly accessible sources (e.g., Wikipedia) and do not contain personally identifiable information beyond what is already publicly disclosed.
The proposed framework, \textsc{QuIeTT}, operates purely on tabular structure and content and does not introduce new data or modify the semantic content of the original tables beyond information preserving  normalization and restructuring. The challenge set introduced in this work is automatically generated from existing tables and manually verified by expert annotators to ensure correctness and faithfulness; no human subjects are involved beyond standard dataset verification, and no personal or sensitive attributes are inferred or labeled.
We do not anticipate any harmful or unethical use of the proposed method. However, as with all large language model-based systems, care should be taken when applying \textsc{QuIeTT} to proprietary, sensitive, or private data, and appropriate data governance and privacy safeguards should be followed. AI tools were used for manuscript and code editing.

\bibliography{custom}

\newpage
\clearpage
\section*{Appendix}
\label{sec:Appendix}

\appendix

\section{Transformation Operators}
\label{app:operators}

QUIETT represents table transformations using a vocabulary of atomic operators that capture common normalization, restructuring, and enrichment patterns observed in semi-structured tables. These operators are deterministic, schema-aware, and composable, enabling reproducible execution across datasets.
Rather than enforcing a closed operator set, QUIETT exposes this vocabulary to the planner as a set of \emph{canonical primitives}. When necessary, the language model may synthesize additional operators, provided they can be expressed as deterministic functions over tabular inputs and conform to the executor interface.

Table~\ref{tab:operator_vocab} summarizes the core operators most frequently used in our implementation.

\section{Resource Usage}
\label{app:cost}

QUIETT incurs language model cost exclusively during a one-time,
\emph{query-independent} preprocessing stage that is executed once per table.
For each table, the pipeline invokes a language model exactly three times:
(i) issue generation, (ii) transformation plan generation, and
(iii) executor code generation. Registered operator execution is deterministic and does not require
any language model calls. LLM-generated code steps are bounded by
the plan structure.

\subsection{Models and Context Limits}

We evaluate the preprocessing pipeline using multiple large language models,
including DeepSeek-V3.1, Qwen-3-80B, Gemini-2.0, Gemini-2.5, and GPT-OSS-120B.
These models support long-context inference, which is necessary to process
table schemas together with representative cell samples within a single prompt.

Table~\ref{tab:model_limits} summarizes the documented context window limits for
each model. Effective throughput and request rates are governed by
provider- and project-level quota policies, which may vary across deployments,
regions, and time.

\begin{table*}[tb]
\centering
\small
\setlength{\tabcolsep}{7pt}
\renewcommand{\arraystretch}{1.15}
\begin{tabular}{l p{0.78\textwidth}}
\toprule
\textbf{Operator} & \textbf{Description} \\
\midrule
add\_row\_id & Adds a unique, stable row identifier column to the table. \\
rename & Renames columns according to a provided name mapping. \\
select & Selects and retains a specified subset of columns. \\
parse\_date\_text & Parses free-text date strings into a normalized date representation. \\
parse\_number & Extracts numeric values (and optional units) from text fields. \\
extract\_regex & Extracts structured groups from a column using a regular expression. \\
derive\_conditional & Creates a new column based on conditional rules applied row-wise. \\
derive\_math & Derives a new column by applying a mathematical expression over existing columns. \\
map\_values & Maps raw values in a column to normalized or canonical values. \\
replace\_value & Replaces exact cell values with specified alternatives. \\
replace\_string & Performs string replacement within column values (optionally regex-based). \\
cast\_column & Casts a column to a specified data type. \\
fillna\_static & Fills missing values using a fixed constant. \\
fillna\_dynamic & Fills missing values using a dynamic rule (\textit{e.g.}, forward fill). \\
combine\_columns & Combines multiple columns into a single column using a separator. \\
trim\_whitespace & Removes leading and trailing whitespace from column values. \\
filter\_rows & Filters rows based on a boolean condition. \\
sort & Sorts rows according to one or more columns. \\
deduplicate\_rows & Removes duplicate rows, optionally using a column subset. \\
keep\_raw\_snapshot & Preserves original raw values in a snapshot column. \\
bin\_numeric & Discretizes numeric values into bins with optional labels. \\
one\_hot & Expands a categorical column into one-hot encoded indicator columns. \\
custom & Applies a user-defined transformation not covered by built-in operators. \\
\bottomrule
\end{tabular}
\caption{Atomic table transformation operators used in the transformation plan.}
\label{tab:operator_vocab}
\end{table*}

\begin{table}[t]
\centering
\small
\setlength{\tabcolsep}{6pt}
\begin{tabular}{ll}
\toprule
\textbf{Model} & \textbf{Context window} \\
\midrule
DeepSeek-V3.1  & Up to 128K tokens \\
Qwen-3-80B     & Up to 128K tokens \\
GPT-OSS-120B   & Up to 131K tokens \\
Gemini-2.0     & Up to $\sim$1,048,576 tokens \\
Gemini-2.5     & Up to $\sim$1,048,576 tokens \\
\bottomrule
\end{tabular}
\caption{Documented context window limits for models used during QUIETT preprocessing. Actual rate limits depend on provider- and project-specific quota configurations.}
\label{tab:model_limits}
\end{table}
\subsection{Token Usage Accounting}

For reproducibility, we log token usage at the granularity of individual tables
and preprocessing stages (issue generation, plan generation, and code
generation). For each language model invocation, we record the number of input
tokens, output tokens, and total tokens consumed.

Total preprocessing token usage over a corpus of tables is computed as:
\[
\text{Tokens}
= \sum_{i=1}^{T}
\sum_{s \in \{\mathrm{issue},\,\mathrm{plan},\,\mathrm{code}\}}
\left( \mathrm{in}_{i,s} + \mathrm{out}_{i,s} \right),
\]
where \(T\) denotes the total number of tables.

\subsection{Amortization Across Queries}
\label{app:amortization}

Because the pipeline preprocesses each table once and caches the result, the marginal
preprocessing token cost per downstream query is zero. All downstream question
answering is performed on the canonical transformed table using an unchanged QA
model, without incurring any additional preprocessing-related language model
cost.

\section{Hyperparameters}
\label{app:hyperparams}

This section reports the inference and execution hyperparameters used in the
QUIETT pipeline. Unless otherwise noted, all LLM calls use deterministic decoding
to improve reproducibility.

\subsection{Language Model Inference Settings}

Table~\ref{tab:llm_hparams} summarizes the decoding and retry configuration used
across pipeline stages. We use greedy decoding (temperature $=0.2$) for all stages.
Maximum output tokens are stage-specific and follow the values used in our
implementation.

\begin{table}[t]
\centering
\small
\setlength{\tabcolsep}{5pt}
\renewcommand{\arraystretch}{1.12}
\begin{tabularx}{\columnwidth}{lX}
\toprule
\textbf{Hyperparameter} & \textbf{Value} \\
\midrule
Decoding & Greedy / temperature $=0.2$ \\
Max output tokens (Step 1: Issue) & 8000 \\
Max output tokens (Step 2: Plan) & 6000 \\
Max output tokens (Step 3: Code) & 1024 \\
Max output tokens (Step 4: Q\&A) & 4096 \\
Stop sequences: & none (generation is post-processed/validated) \\
Retries on rate limit/errors & 3 attempts \\
Backoff schedule & exponential backoff, initial delay $=2.0$s, doubling each retry \\
\bottomrule
\end{tabularx}
\caption{Inference hyperparameters used in the QUIETT pipeline. Step 1-3 are
query-independent preprocessing; Step 4 performs downstream QA on the canonical table.}
\label{tab:llm_hparams}
\end{table}
\subsection{Input Construction and Truncation}

For prompt construction, we serialize each table into markdown that includes the
header and up to 50 rows (uniformly truncated to the first 50 rows when the table
is longer). We do not apply explicit cell-level character truncation; long cell
values are passed through as-is.

\subsection{Executor and Determinism Settings}

A deterministic table executor runs each transformation plan produced in Step~2. Determinism is enforced by fixed operator semantics, stable row referencing (via row identifiers when required), and deterministic implementations for any operation that may reorder rows (\textit{e.g.}, sorting) or drop duplicates (via a
fixed deduplication policy when enabled). The executor introduces no stochasticity.

\subsection{LLM-Generated Code: Safety, Sandboxing, and Determinism}
\label{app:code_safety}

LLM-generated code steps in Stage~2 operate over the in-memory
\texttt{pandas} DataFrame produced by the preceding deterministic
step. The execution scope exposes only \texttt{pandas},
\texttt{numpy}, \texttt{re}, and the operator runtime library,
with no direct filesystem or network access from generated code.
The executor restricts each generated snippet to the declared input
and output columns specified in the plan, and validates that:
(i)~all declared \texttt{writes} columns are present in the output,
(ii)~no all-NaN columns are introduced, and (iii)~row count is
preserved unless the operation legitimately modifies table length
(e.g., \texttt{split\_multi\_value}, \texttt{filter\_rows}, or
\texttt{pivot\_longer}). Snippets that fail any check are rejected
and the model is re-queried with the error appended to the prompt,
up to two additional attempts.

The \texttt{custom} operator serves as an explicit fallback for
transformations that cannot be expressed within the fixed operator
vocabulary. When a plan step declares \texttt{op: custom}, the
executor unconditionally routes to LLM code generation rather than
attempting a registered implementation. All safety and validation
checks described above apply equally to \texttt{custom} steps: the
generated snippet is restricted to the declared input and output
columns, validated against the same three criteria, and re-queried
on failure. 

\section{Human Verification Guidelines and Annotation Rubric}
\label{app:annotation_guidelines}

To assess the correctness, faithfulness, and clarity of the automatically
generated QUIETT challenge set, we conduct a structured human verification
study. This section documents the annotation setup, verification objectives,
and rubric provided to annotators.

\subsection{Annotation Setup}

Each annotator is presented with the following materials:
(i) the \emph{original raw table} drawn from the source dataset,
(ii) a generated natural-language question, and
(iii) the corresponding proposed ground-truth answer.
Annotators perform verification independently and without access to
model predictions or outputs. All judgments are based solely on the
provided table content.

\subsection{Verification Objectives}

Annotators are instructed to evaluate each question-answer pair according
to the following criteria:
\begin{enumerate}
    \item \textbf{Well-formedness}: The question is clearly phrased and free of ambiguity.
    \item \textbf{Answerability}: The question can be answered using only the table content.
    \item \textbf{Answer correctness}: The provided answer is fully supported by the table.
    \item \textbf{Structural reasoning}: The question requires non-trivial reasoning,
    such as aggregation, comparison, filtering, or reasoning over normalized
    or derived fields, rather than simple cell lookup.
\end{enumerate}

Annotators are explicitly instructed not to assess model difficulty,
linguistic style beyond clarity, or subjective preferences.

\subsection{Annotation Guidelines}

The following guidelines are provided to annotators to ensure consistent
judgments:

\begin{itemize}
    \item \textbf{Table faithfulness}: All answers must be derivable exclusively
    from the table content, including normalized or derived columns introduced
    by QUIETT. External knowledge may not be used.

    \item \textbf{Table representation awareness}: Annotators should treat
    the raw table as authoritative. Questions relying on derived or
    normalized fields (e.g., parsed dates or normalized units) are considered
    valid.

    \item \textbf{Ambiguity handling}: If a question admits multiple reasonable
    interpretations or more than one valid answer, it must be marked incorrect.

    \item \textbf{Incomplete answers}: Answers that partially satisfy the question
    (e.g., missing entities in list-valued outputs or incorrect aggregation
    results) are marked incorrect.

    \item \textbf{Assumption prohibition}: Questions requiring assumptions not
    grounded in the table (e.g., inferred semantics, unstated ordering, or
    missing values) are marked incorrect.
\end{itemize}

\subsection{Annotation Rubric}

Each question-answer pair is evaluated using the rubric shown in
Table~\ref{tab:annotation_rubric}. A pair is marked \emph{valid} if and only
if all criteria are satisfied.

\begin{table}[t]
\centering
\resizebox{\columnwidth}{!}{%
\begin{tabular}{l p{8.6cm}}
\toprule
\textbf{Criterion} & \textbf{Description} \\
\midrule
Answer Correctness &
The provided answer exactly matches the value(s) supported by the table. \\
Answerability &
The question can be answered using only the table content. \\
Clarity &
The question is unambiguous and clearly specifies the required reasoning. \\
Structural Reasoning &
The question requires aggregation, comparison, filtering, or reasoning over
derived or normalized fields. \\
\bottomrule
\end{tabular}}
\caption{Annotation rubric used for human verification of the QUIETT challenge set.}
\label{tab:annotation_rubric}
\end{table}

\subsection{Disagreement Analysis}

All annotations are performed independently without discussion or
adjudication during the annotation phase. Inter-annotator agreement is
reported using pairwise Cohen's Kappa and multi-rater agreement metrics
(Fleiss' Kappa and Jaccard coefficient), as described in Section~4.

\subsection{Annotator Qualifications}

All annotators are NLP researchers with prior experience in table-based
question answering, semantic parsing, or structured data reasoning.
We briefed all annotators using written instructions and example annotations before the study.

\subsection{Rationale}

This verification protocol ensures that the QUIETT challenge set evaluates
faithful, well-defined, and structurally grounded reasoning over tables,
rather than exploiting annotation artifacts or superficial patterns. The
high inter-annotator agreement reported in Section~4 indicates strong
consistency in judgments and supports the reliability of the challenge set.


\begin{table*}[tb]
\centering
\scriptsize
\setlength{\aboverulesep}{0.1pt}
\setlength{\belowrulesep}{0.1pt}
\setlength\tabcolsep{10pt}
\setlength{\extrarowheight}{2pt}
\begin{tabular}{c l cc cc cc cc}
\toprule
\rowcolor{quiettgrey}
& & \multicolumn{8}{c}{\textbf{Gemini 2.0}} \\
\cmidrule(lr){3-10}
\multirow{2}{*}{\bf Category} & \multirow{2}{*}{\bf Method}
& \multicolumn{2}{c}{\textbf{WikiTQ}}
& \multicolumn{2}{c}{\textbf{NQ-Table}}
& \multicolumn{2}{c}{\textbf{SequentialQA}}
& \multicolumn{2}{c}{\textbf{HiTab}} \\
\cmidrule(lr){3-4}\cmidrule(lr){5-6}\cmidrule(lr){7-8}\cmidrule(lr){9-10}
& & \bf F1 & \bf EM & \bf F1 & \bf EM & \bf F1 & \bf EM & \bf F1 & \bf EM \\
\midrule
\multirow{8}{*}{\textbf{Direct}}
& CoT           & 61.76 & 60.67 & 68.70 & 67.50 & 62.33 & 59.87 & 76.11 & 72.00 \\
& CoT + SQL     & 44.43 & 44.02 & 43.29 & 41.45 & 50.06 & 46.83 & 52.10 & 50.00 \\
& NormTab       & 55.95 & 52.42 & 51.80 & 49.76 & 47.95 & 42.14 & 56.24 & 53.60 \\
& E5            & 66.60 & 65.90 & 48.40 & 46.34 & 42.75 & 38.18 & 75.83 & 71.40 \\
& EEDP          & 61.80 & 59.30 & 70.31 & 63.81 & 61.10 & 54.27 & 76.53 & 72.00 \\
& Binder        & 60.81 & 57.88 & 62.25 & 54.90 & 21.82 & 20.91 & 55.62 & 53.00 \\
& TableSQLify   & 64.50 & 61.91 & 47.51 & 38.10 & 10.71 & 9.74  & 64.20 & 60.10 \\
& TableReasoner & 60.19 & 57.89 & 56.52 & 53.45 & 59.52 & 51.23 & 60.84 & 54.31 \\
\cmidrule(lr){2-10}
\multirow{4}{*}{\textbf{Agentic}}
& Chain-of-Table & 63.27 & 61.58 & 62.30 & 59.28 & 41.24 & 39.79 & 69.43 & 65.10 \\
& POT            & 52.11 & 50.32 & 61.70 & 52.07 & 56.10 & 52.43 & 64.00 & 61.00 \\
& GraphOTTER     & 55.39 & 51.16 & 69.29 & 65.69 & 50.36 & 49.35 & 64.64 & 61.62 \\
\cmidrule(lr){2-10}
\rowcolor{quiettgreen}
& \textbf{\textsc{QuIeTT}} & \textbf{67.60} & \textbf{66.56} & \textbf{73.25} & \textbf{68.75} & \textbf{70.11} & \textbf{65.69} & \textbf{77.82} & \textbf{76.50} \\
\bottomrule
\end{tabular}
\vspace{-0.75em}
\caption{\small \textbf{Gemini~2.0 results (F1 / EM).} \textsc{QuIeTT} compared against direct prompting methods (CoT, CoT+SQL, NormTab, E5, EEDP, Binder, TableSQLify, TableReasoner) and agentic methods (Chain-of-Table, POT, GraphOTTER) across four benchmarks. Bold indicates best result per column.}
\label{tab:gemini20}
\vspace{-1.0em}
\end{table*}

\begin{table}[t]
\centering
\small
\setlength{\tabcolsep}{4pt}
\renewcommand{\arraystretch}{1.1}
\begin{tabular}{lcccc}
\toprule
\multirow{2}{*}{\textbf{Dataset}} &
\multicolumn{2}{c}{\textbf{Gemini 2.5}} &
\multicolumn{2}{c}{\textbf{GPT-OSS}} \\
\cmidrule(lr){2-3} \cmidrule(lr){4-5}
 & \textbf{F1} & \textbf{EM} & \textbf{F1} & \textbf{EM} \\
\midrule
WikiTQ   & 68.34 & 68.14 & 66.72 & 63.57 \\
NQ-Table& 70.11 & 69.34 & 68.20 & 65.56 \\
\bottomrule
\end{tabular}
\caption{\small \textsc{QuIeTT} performance when a smaller model
(LLaMA-14B Scout) is used for issue generation.}
\label{tab:ablation_slm}
\end{table}
\vspace{-0.5em}

\section{Limitations of Query-Conditioned Transformation}
\label{app:query_conditioned}
Several methods perform table restructuring conditioned on the input query or query set, where schema selection, decomposition, or transformation is guided by the reasoning objective (e.g., Chain-of-Table~\cite{wang2024chainoftable}, TableReasoner~\cite{xiong-etal-2025-teleai}, TabFormer~\cite{dong2025rrdt}, Dater~\cite{10.1145/3539618.3591708}, Binder~\cite{Binder}). While effective for improving efficiency and accuracy on observed queries, this design introduces a fundamental coupling between table representation and query distribution.

First, query-conditioned transformation limits 
generalization to unseen queries. A representation 
tailored to a specific query or narrow query set 
may omit information irrelevant to those queries, 
reducing robustness when new questions arrive, as 
observed in prior table QA and semantic parsing 
work~\cite{yin-etal-2020-tabert, herzig-etal-2020-tapas}.

Second, query-conditioned methods assume query access 
at transformation time, which does not hold in common 
data processing settings where tables are preprocessed 
once and reused across multiple tasks and 
users~\cite{10.1145/2882903.2912574}.

Third, coupling transformation with reasoning 
introduces representation instability: the same table 
yields different transformed views depending on the 
query, making it harder to characterize when and why 
a method fails, since representation quality and 
reasoning behavior are entangled by 
design~\cite{xie-etal-2022-unifiedskg,
pasupat-liang-2015-compositional}.

\textsc{QuIeTT} addresses these limitations by 
converting each table once into a single, 
information-preserving canonical representation 
before any downstream queries arrive, cleanly 
separating representation quality from reasoning 
behavior.
\section{Probe Budget Analysis}
\label{sec:probe_analysis}

QUIETT generates a finite set of synthetic probe queries during issue generation to expose structural deficiencies in the input table. In our implementation we generate $12$-$20$ probe queries per table. This range is treated as a probe budget chosen to balance structural coverage and redundancy.

To empirically validate this design choice, we analyze the number of issues discovered during probing across two benchmark datasets (WikiTQ and NQ-Table) using two backbone models (Gemini~2.5 and GPT-OSS-120B). Across 3,491 table-model pairs, issue discovery naturally concentrates within this range.

Table~\ref{tab:issue_stats} summarizes the distribution of detected issues per table. We observe that $98.5\%$ of WikiTQ tables and $99.4\%$ of NQ-Table tables produce between $11$-$20$ issues, with mean $\approx16.1$ and median $\approx15$.

\begin{table}[t]
\centering
\small
\begin{tabular}{lccc}
\toprule
Dataset & Mean Issues & Median & \% in 11-20 \\
\midrule
WikiTQ & 16.1 & 15 & 98.5\% \\
NQ-Table & 16.1 & 15 & 99.4\% \\
\bottomrule
\end{tabular}
\caption{Issue statistics across datasets demonstrating that most tables naturally fall within the $11$-$20$ issue range.}
\label{tab:issue_stats}
\end{table}

\begin{table}[tb]
\vspace{-0.75em}
\centering
\small
\setlength{\aboverulesep}{0.2pt}
\setlength{\belowrulesep}{0.2pt}
\setlength\tabcolsep{4pt}
\begin{tabular}{l cc cc}
\toprule
\multicolumn{5}{c}{\textbf{Transform model: Gemini 2.5}} \\
\cmidrule(lr){1-5}
\bf QA model $\rightarrow$
& \multicolumn{2}{c}{\bf LLaMA}
& \multicolumn{2}{c}{\bf Mistral} \\
\cmidrule(lr){2-3} \cmidrule(lr){4-5}
& \bf F1 & \bf EM & \bf F1 & \bf EM \\
\midrule
WikiTQ       & 59.87 & 57.50 & 65.12 & 62.23 \\
NQ-Table     & 59.01 & 53.78 & 63.28 & 58.12 \\
SequentialQA & 46.72 & 43.12 & 47.24 & 44.84 \\
HiTab        & 62.30 & 61.20    & 59.61 & 58.40    \\
\midrule
\multicolumn{5}{c}{\textbf{Transform model: Qwen 3}} \\
\cmidrule(lr){1-5}
\bf QA model $\rightarrow$
& \multicolumn{2}{c}{\bf LLaMA}
& \multicolumn{2}{c}{\bf Mistral} \\
\cmidrule(lr){2-3} \cmidrule(lr){4-5}
& \bf F1 & \bf EM & \bf F1 & \bf EM \\
\midrule
WikiTQ       & 53.67 & 50.12 & 63.33 & 59.35 \\
NQ-Table     & 51.60 & 50.70 & 62.56 & 60.23 \\
SequentialQA & 37.70 & 36.50 & 42.81 & 41.12 \\
HiTab        & 46.22 & 45.60    & 51.30 & 47.83    \\
\bottomrule
\end{tabular}
\vspace{-0.75em}
\caption{\small F1 / EM performance across transformation model and QA
model combinations. Stronger transformations consistently improve
downstream QA regardless of QA model size.}
\label{tab:transformation_acl}
\vspace{0.20em}
\end{table}

\begin{figure}[h]
\centering
\includegraphics[width=0.9\linewidth]{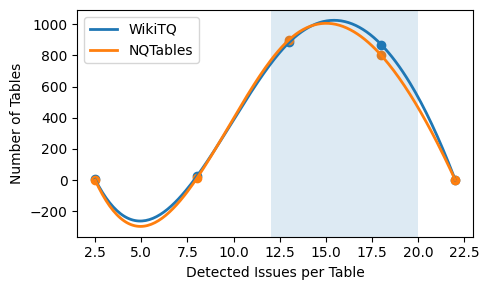}
\caption{Distribution of detected structural issues per table across WikiTQ and NQ-Table. Nearly all tables exhibit between $11$-$20$ issues, supporting the probe budget used by QUIETT.}
\label{fig:issue_distribution}
\end{figure}

Figure~\ref{fig:issue_distribution} visualizes the distribution of detected structural issues per table across datasets. The distribution is concentrated around $\approx16$ issues, indicating that most structural deficiencies are uncovered within this probing regime.

We further observe that probe count correlates strongly with issue discovery ($r\approx0.72$), suggesting that each probe typically surfaces a distinct structural deficiency. At the same time, no tables exceed $20$ detected issues, indicating diminishing returns beyond this range.

Together, these observations suggest that the $12$-$20$ probe range provides sufficient structural coverage while avoiding redundant probing.
\section{Grounding Analysis of Transformed Values}
\label{app:grounding_analysis}

To assess whether \textsc{QuIeTT} preserves source-table content during transformation, we conduct a post-hoc cell-level grounding analysis over transformed tables. For each output value in the canonical table, we determine whether it is (a) an exact string match to a value appearing in the original table or (b) a deterministic derivation of source content. Deterministic derivations include operations such as splitting ``John Doe'' into first and last name columns, extracting year components from date strings, or casting string values to numeric types.

Our analysis shows that 86.0\% of output values are exact string matches to source cells, while the remaining 14.0\% arise from deterministic derivations grounded in the original table. We do not observe unsupported values that lack a source-table justification. These findings provide additional empirical evidence that \textsc{QuIeTT} preserves table content while making structure more explicit for downstream reasoning.

This analysis is intended as supporting evidence for grounding and traceability. It is not an additional runtime constraint of the executor; rather, it is a post-hoc measurement of how transformed values relate to the source table.

We note that information preservation, as defined in 
constraint~(b) of Section~\ref{sec:framework}, refers to grounding 
and provenance rather than strict invertibility. The canonical 
table retains all source content in an accessible, structured form, 
though exact reconstruction of the original surface representation 
is not guaranteed for values that undergo parsing or normalization 
(e.g., date strings converted to ISO format).

\section{Illustrative Examples}
\label{app:examples}

This section presents representative examples highlighting common
failure modes in table-based reasoning and how QUIETT transformations
resolve them.
\subsection{Example 1: Hierarchical Table}
\begin{figure}[h]
\centering
\includegraphics[width=\linewidth]{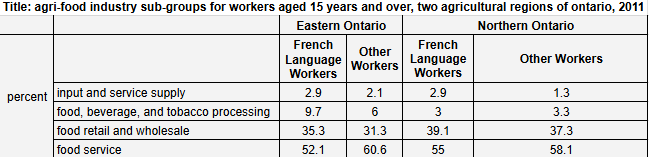}
\caption{Original hierarchical table with multi-level headers for agri-food employment distribution across two Ontario regions (2011).}
\label{fig:raw_table_example}
\end{figure}
\paragraph{Flattened Representation.}

\begin{figure}[H]
\centering
\includegraphics[width=\linewidth]{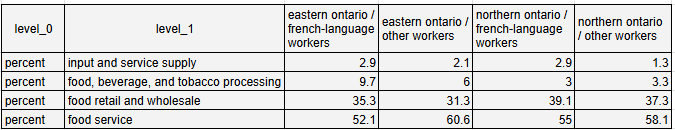}
\caption{Flattened table obtained by collapsing hierarchical column headers into composite columns combining region and worker type.}
\label{fig:flattened_table_example}
\end{figure}

The original table (Figure~\ref{fig:raw_table_example}) contains
hierarchical headers encoding attributes such as region and worker
category. We flatten this structure by collapsing multi-level headers
into composite column names (e.g., ``eastern ontario / french-language
workers''), as shown in Figure~\ref{fig:flattened_table_example}.

While this removes the visual hierarchy, the underlying semantics remain
implicitly encoded in the column names. As a result, models must infer
both structure and meaning from text, often leading to reasoning and SQL
generation errors.
\begin{tcolorbox}[
title=Question and Reasoning Failure,
colback=white,
colframe=black,
breakable,
sharp corners,
boxrule=0.8pt
]

\textbf{Question:}

In Northern Ontario, what proportion of workers in the restaurant and
food services sector were French-language workers?

\vspace{4pt}

\textbf{Generated CoT (Incorrect):}

\begin{quote}
The question refers to the ``food service'' row in
Figure~\ref{fig:flattened_table_example}. For Northern Ontario,
the value for French-language workers is 55.0 and for other workers
is 58.1. To compute the proportion of French-language workers,
we divide 55.0 by (55.0 + 58.1), resulting in 0.486.
\end{quote}

\textbf{Predicted Answer:} 48.6 

\textbf{Gold Answer:} 55.0 

\vspace{4pt}
\textbf{Reasoning Context:}

The query requires interpreting composite column semantics in
Figure~\ref{fig:flattened_table_example}.

\textbf{Issue:}

The reasoning misinterprets the values in
Figure~\ref{fig:flattened_table_example} as counts that can be combined,
whereas they are already percentages representing the distribution of
workers within each category. The correct answer corresponds directly
to the value in the \textit{``northern ontario / french-language workers''}
column for the \textit{``food service''} row (55.0), without requiring
any further computation.

\end{tcolorbox}

\paragraph{Relational Transformation (QUIETT).}

\begin{figure*}[t]
\centering
\includegraphics[width=\textwidth]{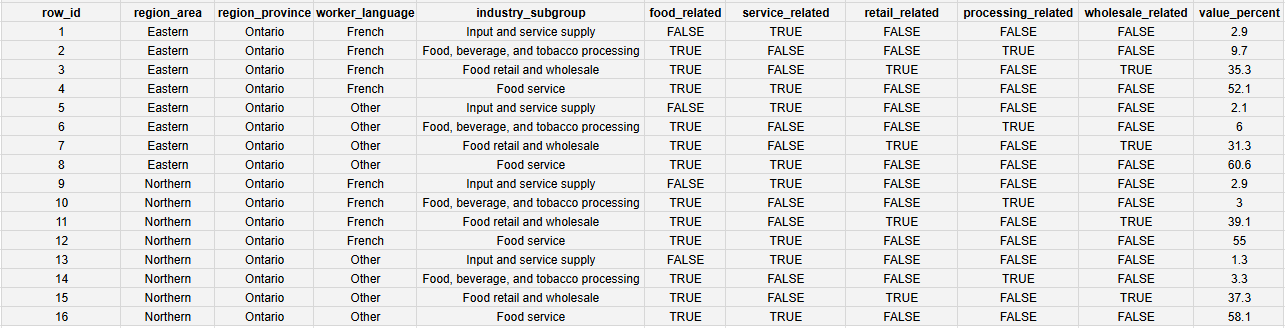}
\caption{Relational table produced by QUIETT after decomposing composite headers into atomic attributes such as region and worker type. Each row represents a single observation suitable for SQL generation.}
\label{fig:transformed_table_example}
\end{figure*}

QUIETT transforms the flattened table into a normalized relational
schema by decomposing composite headers into explicit attributes such as
\textit{region} and \textit{worker type}. Each row represents a single
observation with well-defined fields, as shown in
Figure~\ref{fig:transformed_table_example}.

\begin{tcolorbox}[
title=Correct SQL After Transformation,
colback=white,
colframe=black,
breakable,
sharp corners,
boxrule=0.8pt,
before skip=10pt
]

\textbf{Generated SQL (Correct):}

\begin{lstlisting}[language=SQL, breaklines=true]
SELECT value_percent
FROM transformed_table
WHERE region_area = 'Northern'
  AND region_province = 'Ontario'
  AND industry_subgroup = 'Food service'
  AND worker_language = 'French';
\end{lstlisting}

\textbf{Predicted Answer:} 55.0

\textbf{Gold Answer:} 55.0

\textbf{Observation:}

The transformation (Figure~\ref{fig:transformed_table_example})
eliminates ambiguity by representing attributes such as \textit{region}
and \textit{worker language} as explicit columns. This allows the query
to operate directly on atomic fields, enabling accurate mapping from
natural language to schema and correct retrieval of the expected value.

\end{tcolorbox}

\subsection{Example 2: WebTables}

\begin{figure}[h]
\centering
\includegraphics[width=\linewidth]{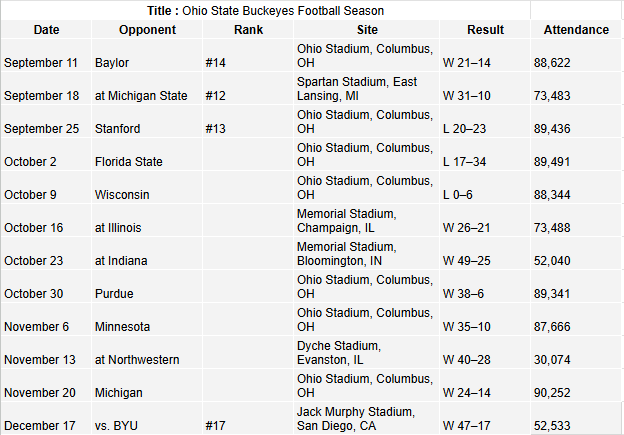}
\caption{Original WebTable containing Ohio State Buckeyes football season data. The table is already flat but contains implicit structure such as game location, ranking, and match outcomes embedded in textual fields.}
\label{fig:webtable_raw}
\end{figure}

\paragraph{Initial Representation.}
The original WebTable (Figure~\ref{fig:webtable_raw}) presents Ohio State
Buckeyes football season data in a flat format with columns such as
\textit{date}, \textit{opponent}, \textit{rank}, \textit{site},
\textit{result}, and \textit{attendance}. Although the structure is not
hierarchical, multiple semantic attributes are embedded within individual
columns.

For example, the \textit{result} column in Figure~\ref{fig:webtable_raw}
combines outcome and score (e.g., ``W 21-14''), while the \textit{site}
column merges stadium and location, and the \textit{rank} column uses
textual annotations (e.g., ``\#14''). These implicit encodings, as shown
in Figure~\ref{fig:webtable_raw}, require string parsing and can lead to
errors in SQL generation.
\begin{tcolorbox}[
title=Question and Reasoning Failure,
colback=white,
colframe=black,
breakable,
sharp corners,
boxrule=0.8pt
]

\textbf{Question:}

Which opponent had the largest amount of points scored?

\textbf{Generated CoT (Incorrect):}

\begin{quote}
The question asks to find the opponent against whom the Buckeyes scored
the largest number of points. Looking at the \textit{result} column in
Figure~\ref{fig:webtable_raw}, we identify the largest number to the left
of the dash. The maximum value is 49, which corresponds to Indiana.
\end{quote}

\textbf{Predicted Answer:} Indiana 

\textbf{Gold Answer:} Florida State

\vspace{4pt}
\textbf{Reasoning Context:}

The query requires extracting structured information from the
\textit{result} column in Figure~\ref{fig:webtable_raw}.

\textbf{Issue:}

The reasoning incorrectly interprets the \textit{result} column in
Figure~\ref{fig:webtable_raw} by assuming that the left-hand value always
represents the Buckeyes' score. However, the ordering of scores varies
depending on whether the game was played at home or away. As a result,
the model extracts an incorrect maximum value (49) and selects the wrong
opponent.

\end{tcolorbox}
\paragraph{Relational Transformation (QUIETT).}

\begin{figure*}[t]
\centering
\includegraphics[width=\textwidth]{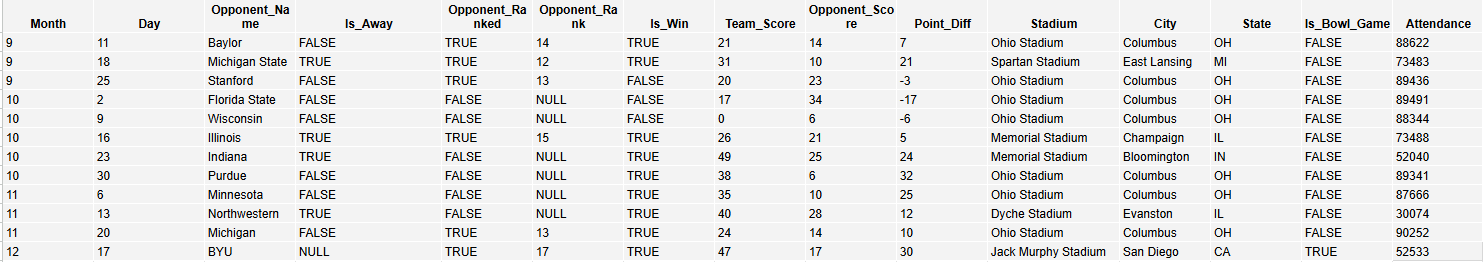}
\caption{QUIETT-transformed relational representation of the original WebTable (Figure~\ref{fig:webtable_raw}). Composite fields such as \textit{result} and \textit{site} are decomposed into atomic attributes including \textit{Is\_Win}, \textit{Team\_Score}, \textit{Opponent\_Score}, \textit{Stadium}, \textit{City}, and \textit{State}. The resulting schema enables direct and unambiguous SQL querying.}
\label{fig:webtable_transformed}
\end{figure*}

The transformed table (Figure~\ref{fig:webtable_transformed}) converts
the original WebTable (Figure~\ref{fig:webtable_raw}) into a normalized
schema by decomposing composite fields into atomic attributes. In
particular, the \textit{result} column is split into
\textit{Is\_Win}, \textit{Team\_Score}, and \textit{Opponent\_Score},
and the \textit{site} column into \textit{Stadium}, \textit{City}, and
\textit{State}.

This representation, as shown in
Figure~\ref{fig:webtable_transformed}, removes ambiguity in textual
encodings and enables direct, reliable SQL querying over structured
fields.

\begin{tcolorbox}[
title=Correct SQL After Transformation,
colback=white,
colframe=black,
breakable,
sharp corners,
boxrule=0.8pt
]

\textbf{Generated SQL (Correct):}

\begin{lstlisting}[language=SQL]
SELECT Opponent_Name
FROM games
ORDER BY Opponent_Score DESC
LIMIT 1;
\end{lstlisting}

\textbf{Predicted Answer:} Florida State

\textbf{Gold Answer:} Florida State

\vspace{4pt}

\textbf{Observation:}

Using the transformed table (Figure~\ref{fig:webtable_transformed}),
the query directly operates on the structured attribute
\textit{Opponent\_Score}, enabling correct identification of the
maximum value without requiring string parsing.

\end{tcolorbox}

\section{Prompt Templates}
\label{app:prompts}

For readability, we summarize the fixed stage-wise prompts here. The released code contains the exact prompt strings used in all experiments.

\subsection{Issue Generation (Step~1)}
\label{app:issue_gen}

\begin{tcolorbox}[
  title=Issue Generation Prompt,
  colback=white,
  colframe=black,
  breakable,
  sharp corners,
  boxrule=0.8pt
]
\begin{lstlisting}[basicstyle=\ttfamily\scriptsize,breaklines=true,
breakatwhitespace=false,columns=fullflexible]
You are generating analysis questions from a RAW table.

Table Title: {table_title}
Column Descriptions: {column_descriptions}
Table (Markdown): {raw_markdown_table}

Objective:
- Produce 12-20 diverse, realistic questions that expose structural
  or semantic deficiencies in the raw table.
- Questions should not be reliably answerable from this RAW table
  without cleaning, structuring, or transforming the data.

Output format:
- Return ONLY a JSON ARRAY (no prose, no markdown, no backticks).
- Each item MUST be a JSON OBJECT with these exact keys:
  - "qid": string like "Q1", "Q2", ... (contiguous, no gaps)
  - "text": natural language question
  - "depends_on": list of raw column names, or ["unknown"] if unclear
  - "requires": list of data-prep needs
  - "failure_reason": short phrase on why naive analysis fails

- Return ONLY raw JSON. No text before or after the array.
- All strings must use double quotes. No trailing commas.
\end{lstlisting}
\end{tcolorbox}

\paragraph{Issue Aggregation.}
Step-1 questions are post-processed by grouping on identical
\texttt{failure\_reason} strings. Each unique failure reason becomes
one \emph{issue} (\texttt{issue\_id}, \texttt{description}, union of
referenced \texttt{cols}, and the list of
\texttt{blocking\_questions}). The resulting \texttt{issues} array
is the input to Step~2.

\subsection{Plan Generation (Step~2)}
\label{app:plan_gen}

\begin{tcolorbox}[
  title=Plan Generation Prompt,
  colback=white,
  colframe=black,
  breakable,
  sharp corners,
  boxrule=0.8pt,
  width=\columnwidth
]
\begin{lstlisting}[basicstyle=\ttfamily\scriptsize,breaklines=true,
breakatwhitespace=false,columns=fullflexible]
You are a transformation planner for arbitrary tabular data.
You NEVER write code. Output only a MACHINE-EXECUTABLE PLAN in JSON.

Inputs:
- TABLE_TITLE        : short description of the table.
- COLUMN_DESCRIPTIONS: optional text describing columns.
- TABLE_PREVIEW      : first 3 rows in Markdown.
- ISSUES_JSON        : issues detected in Step 1.

Design a transformation plan that:
  1. FIXES ALL ISSUES from ISSUES_JSON
  2. OPENS UP THE TABLE with derived columns
  3. PRESERVES original information without redundancy

Output ONE JSON OBJECT:
{
  "table_id": string,
  "strategy": string,
  "steps": [
    {
      "step_id":      string,
      "op":           string,
      "description":  string,
      "reads":        [string],
      "writes":       [string],
      "params":       { ... },
      "fixes_issues": [string],
      "depends_on":   [string]
    }
  ],
  "final_output": {
    "primary_key": [string],
    "columns": [
      {
        "name": string,
        "role": "canonical" | "derived" | "helper" | "raw_snapshot"
      }
    ]
  }
}

Column roles:
  canonical   : original column retained in final table
  derived     : new column created; always in final table
  helper      : intermediate column; dropped via select
  raw_snapshot: original column dropped from final table

Use only operators from the fixed operator vocabulary
(Table~\ref{tab:operator_vocab}). Each op must match a registry
entry and its params must satisfy required-parameter constraints.

CRITICAL CONSTRAINTS:

1. derive_conditional - use EXACT structure:
   { "conditions": [{"condition": "col contains 'v'", "value": true}],
     "default": false }

2. derive_math:
   "len(col)", "year(col)", "col_a + col_b"

3. extract_regex / parse_number - include commas in numeric patterns:
   "[0-9,]+" not "[0-9]+"

4. Row count may increase only when split_multi_value with
   explode=true is explicitly declared in the plan strategy.

COLUMN RETENTION:
  KEEP (canonical) when derived columns capture only PART of the info.
  DROP (raw_snapshot) when derived columns capture ALL info, e.g.:
    "W 27-7"   -> is_win + score_for + score_against -> raw_snapshot
    "45.2%"    -> pct_numeric                        -> raw_snapshot
    "1990-2005"-> start_year + end_year              -> raw_snapshot

WORKED EXAMPLES:

EXAMPLE A - Athletics Table (numeric, ordinal, boolean):
  Plan: add_row_id -> parse_number(Best) -> map_values(Position) ->
        cast_column(pos_numeric) -> derive_conditional(is_podium) ->
        select(...)

EXAMPLE B - Race Results with Multi-Driver Rows (explode):
  Plan: split_multi_value(Co-Drivers, explode=true) ->
        trim_whitespace -> add_row_id ->
        derive_conditional(is_classified) -> select(...)

ANTI-PATTERNS (NEVER DO):
  1. Writing raw "W"/"L" instead of is_win boolean.
  2. cast_column without out= parameter.
  3. Skipping the final select step.
  4. Exploding a notes or description column.

MANDATORY: Every plan MUST end with a select step containing:
  _row_id, all derived columns, all canonical columns.
  NO raw_snapshot or helper columns.

Return ONLY the JSON object. No markdown, no backticks, no comments.
\end{lstlisting}
\end{tcolorbox}

\paragraph{Plan Validation.}
The emitted plan is validated against a fixed operator schema before
execution. Unknown operators, missing required parameters, undefined
\texttt{reads}, dangling \texttt{depends\_on}, or duplicate
\texttt{writes} are all rejected. A failed validation triggers a
single regeneration attempt with the validator's error list appended
to the prompt.

\subsection{Code Generation (Step~3)}
\label{app:code_gen}

\begin{tcolorbox}[
  title=Code Generation Prompt,
  colback=white,
  colframe=black,
  breakable,
  sharp corners,
  boxrule=0.8pt,
  width=\columnwidth
]
\begin{lstlisting}[basicstyle=\ttfamily\scriptsize,breaklines=true,
breakatwhitespace=false,columns=fullflexible]
You are a Python Data Engineer. Write a snippet to transform
DataFrame `df`.

CONTEXT:
- Current columns: {columns}
- DataFrame shape: {n_rows} rows x {n_cols} columns
- Sample data (first 3 rows of input columns): {sample_rows}

TASK: {task_description}
OPERATION: {operation}
Parameters: {parameters}
Input columns: {input_columns}
Output columns: {output_columns}

REQUIREMENTS:
1. All output columns MUST be created.
2. On extraction/parsing failure, fall back to the original value.
3. Do NOT create all-NaN columns.
4. Handle missing data gracefully.
5. Do NOT drop rows unless explicitly required.

EXAMPLE PATTERNS:
  df['new'] = df['src'].str.extract(r'pattern')[0]
  df['new'] = df['new'].fillna(df['src'])  # fallback
  df['parsed'] = pd.to_datetime(df['date'], errors='coerce')

Return ONLY valid Python code. Assume df, pd, np, re exist.
\end{lstlisting}
\end{tcolorbox}

\paragraph{Execution Loop.}
For every step in the plan, Step~3 first attempts deterministic
execution via the operator library. Only when the operation is
\texttt{custom} or the deterministic call raises an exception does
the system fall back to LLM code generation. The generated snippet
is checked for: (i)~all declared \texttt{writes} columns existing,
(ii)~no all-NaN columns, and (iii)~row-count preservation unless
\texttt{split\_multi\_value} with \texttt{explode=true} is declared.
Validation failures are appended to the prompt and the model is
re-queried up to two additional times.

\subsection{SQL Query Generation for Q/A}
\label{app:qa_gen}

\begin{tcolorbox}[
  title= Question Answering Prompt,
  colback=white,
  colframe=black,
  breakable,
  sharp corners,
  boxrule=0.8pt,
  width=\columnwidth
]
\begin{lstlisting}[
  basicstyle=\ttfamily\scriptsize,
  breaklines=true,
  breakatwhitespace=false,
  columns=fullflexible
]
You are answering a question about a structured table. Analyze the table and find the answer.

## TABLE: {table_name}

### Columns:
{column_description}

### Data:
{transformed_table}

-

## QUESTION: {question}

-


1. **Interpret the question** - Identify what information is being requested

2. **Identify relevant columns** - Determine which column(s) contain the needed data

3. **Determine filtering criteria** - Identify any row-level conditions (e.g., specific entity, date, value)

4. **Identify required operation** - Determine if aggregation is needed:
   - Counting: COUNT
   - Maximum/minimum: MAX/MIN or ORDER BY with LIMIT
   - Summation: SUM
   - Lookup: Direct selection

5. **Extract the answer** - Locate the answer in the table data

<reasoning>
[Step-by-step analysis]
</reasoning>

<answer>
[Extracted answer value]
</answer>

<sql_plan>
SELECT: [target column]
FROM: {table_name}
WHERE: [filter condition if applicable]
ORDER BY: [ordering if applicable]
AGGREGATION: [function if applicable]
</sql_plan>





\end{lstlisting}
\end{tcolorbox}

\end{document}